\def\year{2021}\relax
\documentclass[letterpaper]{article} 
\usepackage{aaai21}  
\usepackage{times}  
\usepackage{helvet} 
\usepackage{courier}  
\usepackage[hyphens]{url}  
\usepackage{graphicx} 
\usepackage{natbib}  
\usepackage{caption} 
\usepackage{pdfpages}

\usepackage{booktabs}       
\usepackage{amsfonts}       
\usepackage{nicefrac}       
\usepackage{microtype}      
\usepackage{subfigure}
\usepackage{graphicx}
\usepackage{multirow}
\usepackage{float} 
\usepackage{color}
\usepackage{wrapfig}
\usepackage{setspace}
\usepackage{appendix}
\usepackage{algorithm}
\usepackage{algorithmic}
\usepackage{amsmath}

\newcommand{\myparagraph}[1]{\noindent\textbf{#1}}
\newcommand{\ie}{\textit{i.e.}}
\newcommand{\eg}{\textit{e.g.}}
\newcommand{\softmax}{\mathrm{softmax}}

\title{Counterfactual Variable Control for Robust and Interpretable Question Answering}
\author {

        Sicheng Yu\textsuperscript{\rm 1},
        Yulei Niu\textsuperscript{\rm 2},
        Shuohang Wang\textsuperscript{\rm 3},
        Jing Jiang\textsuperscript{\rm 1},
        Qianru Sun\textsuperscript{\rm 1}\\
}
\affiliations {
    \textsuperscript{\rm 1} Singapore Management University,
    \textsuperscript{\rm 2} Nanyang Technological University, 
    \textsuperscript{\rm 3} Microsoft Dynamics 365 AI Research
    scyu.2018@phdcs.smu.edu.sg, yn.yuleiniu@gmail.com, shuowa@microsoft.com,
    \{jingjiang, qianrusun\}@smu.edu.sg
}
\begin{document}
\maketitle

\begin{abstract}

Deep neural network based question answering (QA) models are neither robust nor explainable in many cases.
For example, 
a multiple-choice QA model, tested without any input of \emph{question}, is surprisingly ``capable'' to predict the most of correct options.
In this paper, we inspect such spurious 
``capability'' of QA models using causal inference. We find the crux is the shortcut correlation, \eg, unrobust word alignment between \emph{passage} and \emph{options} learned by the models.
We propose a novel approach called Counterfactual Variable Control (CVC) that explicitly mitigates any shortcut correlation and preserves the comprehensive reasoning for robust QA.
Specifically, we leverage multi-branch architecture~\cite{cadene2019rubi} that allows us to disentangle robust and shortcut correlations in the training process of QA. 
We then conduct two novel CVC inference methods (on trained models) to capture the effect of comprehensive reasoning as the final prediction.
For evaluation, we conduct extensive experiments using two BERT backbones on both multi-choice and span-extraction QA benchmarks. 
The results show that our CVC achieves high robustness against a variety of adversarial attacks in QA while maintaining good interpretation ability\footnote{Our code is publicly available on GitHub: \url{https://github.com/PluviophileYU/CVC-QA}}.
\end{abstract}

\section{Introduction}
\label{sec_intro}

Recently, the error rates
on the multiple-choice question answering (MCQA) and span-extraction question answering (SEQA) benchmarks were smashed overnight by large-scale pre-trained models such as BERT~\cite{devlin2019bert}, XLNet~\cite{yang2019xlnet}, RoBERTa~\cite{liu2019roberta} and Megatron-LM~\cite{shoeybi2019megatron}.
Impressively, using Megatron-LM achieved less than $10\%$ error on RACE~\cite{lai2017race}.
However, top-performing models often lack interpretability~\cite{feng2018pathologies, kaushik2018much}, nor are they robust to adversarial attacks~\cite{ribeiro2018semantically, SzegedyZSBEGF13, wallace2019universal}.
For example, adding one more question mark to the input \emph{question}, 
which is
a simple adversarial attack, may decrease the performance of a QA model~\cite{ribeiro2018semantically}.
This kind of vulnerability will raise security concerns when the model is deployed in real systems, \eg, intelligent shopping assistant.
It is thus desirable to figure out why this happens and how to improve the robustness of the model.

\begin{figure*}
    \centering
    \includegraphics[width=0.95\textwidth]{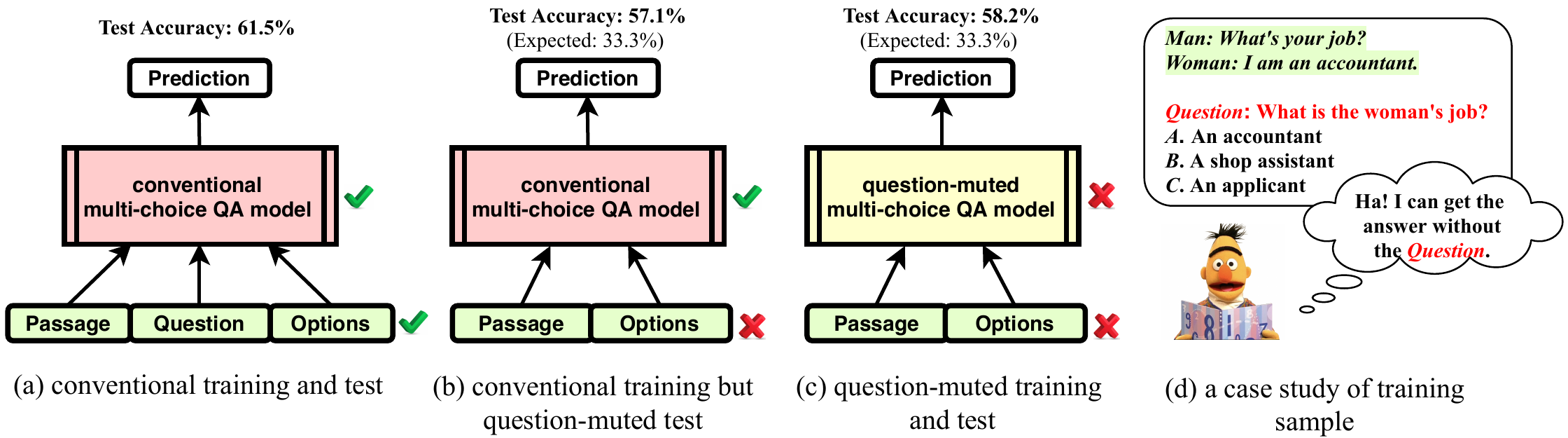}
    \caption{We observe multi-choice QA models are ``capable'' to answer a question without any \emph{question} data in input (question-muted) during test (b), or during both training and test (c).
    We conduct these experiments using the BERT-base model~\cite{devlin2019bert} on the multi-choice QA benchmark DREAM~\cite{sun2019dream}. 
    (a) shows the normal case for reference. (d) show a training sample on DREAM. 
    }
    \label{fig_tisser}
\end{figure*}

In this paper, we carefully inspect the training and test processes for QA models. We find the mentioned vulnerability is caused by the fact that
%
the model often overfits to the \emph{correlation} in training.
To illustrate this, we show some 
example results of BERT-base MCQA model~\cite{devlin2019bert} in Figure~\ref{fig_tisser}.
Specifically,
(a) is the conventional result as a reference. (b) and (c) are special results as there is \emph{no question} in the model input.
It is surprising that \emph{no question} during test as in (b), or during both training and test as in (c), 
has very little performance drop, \ie, less than $5$ percentage point in accuracy.
Our 
hypothesis
is that the BERT-base MCQA model uses a huge amount of network parameters to brutally learn the \emph{shortcut correlation} between the \emph{no question} input (\emph{passage} and \emph{options} only) and the ground truth \emph{answer}. 
We give an example in Figure~\ref{fig_tisser}~(d) to show that this \emph{shortcut} could be achieved by aligning words between the \emph{passage} and the \emph{options}.
Can we just conclude that \emph{questions} have little effect on \emph{answers}? We must say no, as we understand this totally violates our common sense about the causality of QA --- \emph{the question causes the answer}. 
Similarly, we may ask what effect the \emph{passage} has.

\begin{figure}[ht]
\centering
    \includegraphics[width=0.38\textwidth]{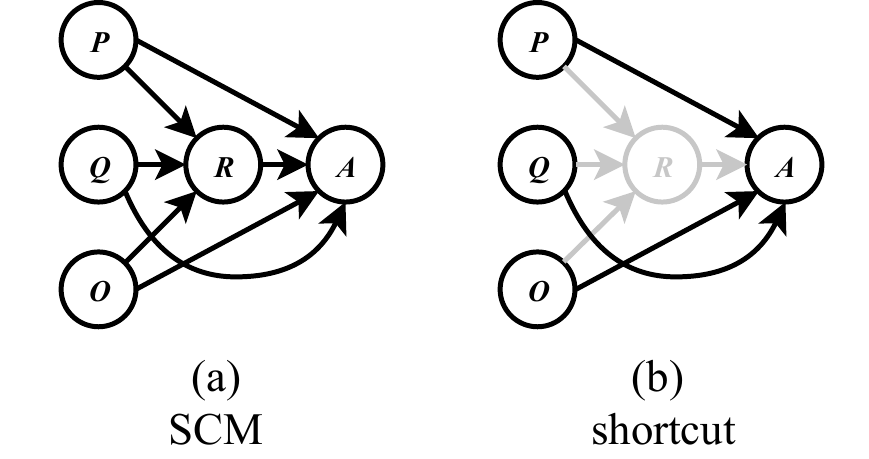}
\caption{
The SCM of MCQA.
\textit{P} is for \emph{passage}, \textit{Q} for \emph{question},
\textit{O} for \emph{options} and \textit{A} for \emph{answer}.
Particularly, \textit{R} denotes the
comprehensive \emph{reasoning}.
}
\label{fig_human_machine}
\end{figure}

Driven by these questions in mind, we try to figure out the overall causality of QA in this paper, based on causal inference~\cite{pearl2009causal, pearl2018book}.
We begin by analyzing the causal relationships among QA variables, \ie, associating any two variables conditional on the causal effect from one on another.
%
Thanks to the representation methods of causality~\cite{pearl2009causal, qi2019two, tang2020unbiased,zhang2020causal}, we can represent these QA relationships using the Structural Causal Model (SCM).
In Figure~\ref{fig_human_machine}~(a), we show the SCM we build for MCQA as an example.
Each node denotes a variable, \eg, \textit{Q} for \emph{question}, and the arrow between two nodes denotes a causal relationship, \eg, $\textit{Q}\!\rightarrow\!\textit{A}$ represents \emph{question causes answer}. 
It is worth mentioning that $\textit{R}$ stands for the comprehensive \emph{reasoning} among \emph{passage}, \emph{question} and \emph{options}, which is expected in robust QA.


Based on the SCM, it becomes obvious that not only comprehensive \emph{reasoning} but also shortcut correlations have effects on the prediction results of QA models. 
For example, as highlighted in Figure~\ref{fig_human_machine}~(b), \textit{P} and \textit{O} can still reach \textit{A} without \textit{Q} (the success rate is $24\%$ higher than random guess, as shown in Figure~\ref{fig_tisser}~(b)).
These shortcut correlations are ``distractors'' against our goal of robust QA (i.e., the prediction should be caused by only the comprehensive \emph{reasoning}).
To alleviate shortcuts, we propose a novel approach called Counterfactual Variable Control (CVC) based on causality theory. 
CVC in essence includes \emph{counterfactual analysis}~\cite{pearl2009causal,pearl2018book,pearl2001direct} and \emph{variable control}.
The first one imagines a counterfactual world~\cite{Roese1997cfthinking} where the controlled variable (\eg, \textit{Q}) had existed to derive its subsequent variables (\eg, \textit{R}).
This avoids any interference on uncontrolled variables that are not input (\eg, \textit{R}).
The second one enables us to explicitly separate comprehensive reasoning and shortcut correlations by variable control, \eg, muting \textit{Q} in Figure~\ref{fig_tisser} (b) and (c).
To implement CVC in deep models, we leverage a multi-branch neural network architecture ~\cite{cadene2019rubi} composed of a robust (comprehensive reasoning) branch and several shortcut branches.
We highlight that CVC training exactly follows the conventional supervised training, while CVC testing is based on \emph{counterfactual analysis} to capture the indirect effects (\ie, the predictions from comprehensive reasoning only) for robust QA.
In experiments, we validate the efficiency and generality of the proposed CVC approach 
by 
using different backbone networks such as BERT-base and BERT-large~\cite{devlin2019bert}
and 
experimenting on four QA benchmarks.

Our contributions thus include (i) an overall causality analysis using SCM for QA; 
(ii) a novel CVC approach to mitigate the shortcut correlations while preserving the robust reasoning in QA;
(iii) plugging CVC in different deep backbones and evaluate it on large-scale QA benchmarks.

\section{Related Work}
\label{sec_related}

\myparagraph{Robustness in NLP.}
Many recent works generate adversarial examples to augment the training data such as to make the trained more robust against adversarial attacks~\cite{ribeiro2018semantically,liurobust,jia2017adversarial,wang2018robust}.
They achieve a fairly good performance but their problems are obvious. 
First, they need to be aware of the prior knowledge of the coming adversarial attack, \ie, ``in what way to generate adversarial examples'', which is often not available in real applications.
Second, their model performance strongly relies on the quality of adversarial examples as well as the training hyperparameters (\eg, augmentation ratios).
Alternative methods (for robustness in NLP) include using advanced regularizer~\cite{yeh2019qainfomax}, training loss \cite{jia2019certified,huang2019achieving}, sample filtering \cite{yaghoobzadeh2019robust,bras2020adversarial} 
and model ensembling \cite{clark2019don,cadene2019rubi,he2019unlearn,utama2020mind}. However, it is uncertain why and how these methods can train the QA model to do robust inference.
\textbf{The relationship and difference between them and our approach are as follows.} 
In terms of the detailed implementation, our CVC is related to model ensembling.
While, we want to highlight that our systematical and explainable causal formulation for QA is the main contribution that potentially opens a principled direction to understanding the real problems of existing QA models and finding out solutions.
So, the mentioned model ensembling works can be regarded as implementation-level examples under our formulation.

\myparagraph{Adversarial Attacks in QA.}
Adversarial attack 
is an indirect way to
evaluate the robustness of machine models~\cite{zhang2020adversarial}.
The works on NLP adversarial attack can be 
roughly divided into two 
categories:
word/sentence-level attack and character-level attack. 
The first category includes text paraphrasing~\cite{ribeiro2018semantically,zhang2019paws,iyyer2018adversarial} 
and word substitution~\cite{ren2019generating,zhang2019generating,alzantot2018generating},
and
the second category is mainly based on character-wise perturbation~\cite{ebrahimi2018adversarial,ebrahimi2018hotflip}.
Our used adversarial attack methods belong to the first category.
Besides, most of previous works propose ``how to attack'' but rarely mention ``how to defend''.
Our work considers both, and importantly, our proposed defending approach is validated robust against unseen attacks (in our experiments).

\myparagraph{Causal Inference in Deep Learning.}
Causal inference~\cite{pearl2009causal,pearl2018book,neuberg2003causality} is based on the causal assumption make in each specific task, \eg, QA task in this paper.
It has been widely applied to epidemiology~\cite{rothman2005causation}, computer science~\cite{van2011targeted}, and social science~\cite{steel2004social}. 
Recently, it is incorporated
in a variety of deep learning applications such as image classification~\cite{goyal2019counterfactual}, image parsing~\cite{zhang2020causal}, few-shot learning~\cite{yue2020interventional}, long-tail problems~\cite{tang2020long}, pre-training~\cite{wang2020visual}, scene graph generation~\cite{tang2020unbiased} and vision-language tasks~\cite{qi2019two,yang2020deconfounded,chen2020counterfactual,niu2020counterfactual,abbasnejad2020counterfactual}. 
In NLP, \cite{kaushik2020,lawrence2018improving} %
propose causal inference methods used for SNLI~\cite{bowman2015large} and semantic parsing.
In contrast, our work makes the first trial to improve the robustness of QA models.


\section{Counterfactual Variable Control (CVC)}
\label{sec_method_part1}

\textbf{Notation.} We use MCQA (with its SCM given in Figure~\ref{fig_human_machine}) as a study case of QA tasks, and introduce the notation for CVC.
Basically, we use uppercase letters to
denote MCQA variables (\eg, $\textit{Q}$ for \emph{question}) and lowercase letters
for the value of a variable (\eg, $q$ for a specific question). 
We also introduce the notation of counterfactual values of variables, \ie, the imagined values as if the ancestor variables had existed (uncontrolled) in a counterfactual world
~\cite{pearl2009causal,tang2020unbiased,pearl2001direct,Roese1997cfthinking}.
We highlight that our overall notation is general and imposes no constraints on the detailed implementation of QA models.
Based on the input variables (with their normal or counterfactual values), we define the notation for the two cases of model prediction: Normal Prediction and Counterfactual Prediction.

\noindent\textbf{Normal Prediction (NP)} means the model makes a normal prediction according to realistic input values.
We use the function format $Y(X=x)$, abbreviated as $Y_x$, to represent the effect of $X=x$ on $Y$.
We use this format to formulate any path on the SCM, so we can derive the formulation of normal prediction $\textit{A}$ as:
\begin{equation}
        \textit{A}_{p,q,o,r}\!=\!\textit{A}(\textit{P}\!=\!p,\textit{Q}\!=\!q,\textit{O}\!=\!o,\textit{R}\!=\!r),
    \label{e1}
\end{equation}
where $r\!=\!\textit{R}(\textit{P}\!=\!p, \textit{Q}\!=\!q, \textit{O}\!=\!o)$, and $\textit{A}_{p,q,o,r}$ denotes the inference logits of the model.
If all input data are controlled (\eg, by muting their values as null $^*$), we have:
\begin{equation}
        \textit{A}_{p^*,q^*,o^*,r^*}\!=\!\textit{A}(\textit{P}\!=\!p^*,\textit{Q}\!=\!q^*,\textit{O}\!=\!o^*,\textit{R}\!=\!r^*),
    \label{e2}
\end{equation}
where $r^*\!=\!\textit{R}(\textit{P}\!=\!p^*, \textit{Q}\!=\!q^*, \textit{O}\!=\!o^*)$, and $\textit{A}_{p^*,q^*,o^*,r^*}$ is the inference logits of the model with null values of input variables (denoted as $p^*,q^*,o^*$).

\noindent\textbf{Counterfactual Prediction (CP)} means the model makes the prediction when some variables are controlled but the others are assigned with counterfactual values by \emph{imagining a counterfactual world of no variable control}.
For example, when we control $\textit{Q}$ with its input value set to null (denoted as $q^*$), we can assign $\textit{Q}$'s subsequent variable $\textit{R}$ with a counterfactual value $r$ by imagining a world where $q$ had existed and had derived $r$ as its normal value $\textit{R}(\textit{P}\!=\!p, \textit{Q}\!=\!q, \textit{O}\!=\!o)$ (the input of the other two variables are $\textit{Q}\!=\!q, \textit{O}\!=\!o$). This is a key operation in the \emph{counterfactual analysis} \cite{pearl2009causal,pearl2018book,pearl2001direct}.

To conduct CVC inference, we propose two variants of counterfactual control: (i) control only input variables; and (ii) control only mediator variable.
For (i), we can formulate $\textit{A}$ as:
\begin{equation}
    \textit{A}_{p^*,q^*,o^*,r}\!=\!\textit{A}(\textit{P}\!=\!p^*,\textit{Q}\!=\!q^*,\textit{O}\!=\!o^*,\textit{R}\!=\!r),
    \label{e1-1}
\end{equation}
where $p^*,q^*,o^*$ denote null (variables are muted).
Similarly, for (ii), we have: 
\begin{equation}
    \textit{A}_{p,q,o,r^*}\!=\!\textit{A}(\textit{P}\!=\!p,\textit{Q}\!=\!q,\textit{O}\!=\!o,\textit{R}\!=\!r^*),
    \label{e1-2}
\end{equation}
where $\textit{R}$ is the only muted variable.

\noindent\textbf{Using NP and CP for CVC Inference.}
The idea of CVC is to preserve only the robust prediction which is derived by comprehensive reasoning rather than any shortcut correlations.
Thanks to the theory of causality~\cite{morgan2015counterfactuals}, this CVC can be realized by computing the difference between the normal prediction (NP) and the counterfactual prediction (CP).
An intuitive interpretation of this computation is that the importance of a variable can be indirectly revealed by generating the results when imagining this variable had not existed (and then comparing to the real results). If the result difference is not significant, it means this variable is useless, otherwise this variable is important and its pure contribution to the results is exactly the computed difference.
As in our case, such imagination can be applied on either input (\eg, \textit{Q}) or mediator variables (\eg, \textit{R}), so there are two realizations of CVC respectively corresponding to them.

\noindent\textbf{CVC on Input Variables (CVC-IV)}
is derived as:
\begin{equation}
    \mathrm{CVC\mbox{-}IV}=\textit{A}_{p^*,q^*,o^*,r}-\textit{A}_{p^*,q^*,o^*,r^*}
    \label{cvc-iv}
\end{equation}
where particularly in $\textit{A}_{p^*,q^*,o^*,r}$ input variables are controlled (to be null) while the mediator variable uses its counterfactual value (generated by imaging a counterfactual world where there had no control on input variables).

\noindent\textbf{CVC on Mediator Variable (CVC-MV)}
is derived as:
\begin{equation}
    \mathrm{CVC\mbox{-}MV}=\textit{A}_{p,q,o,r}-\textit{A}_{p,q,o,r^*}, 
    \label{cvc-mv}
\end{equation}
where in $\textit{A}_{p,q,o,r^*}$ input variables use observed values while the mediator variable is controlled (\ie, by imagining a counterfactual world where all inputs were set to null).

Both CVC-IV and CVC-MV aim to capture the causal effect of comprehensive \textit{reasoning} in QA. 
Their difference lies in \emph{on which variables to apply the control}. In CVC-IV, it is on the inputs variables (which removes any shortcut correlations), while in CVC-MV, it is on the mediator variable (which preserves only the effect of comprehensive \emph{reasoning} after the subtraction). The idea of CVC-IV is more direct than that of CVC-MV. 
We will show in experiment that they perform differently in different experimental settings of QA.
\section{The Implementation of CVC}
\label{sec_method_part2}
\begin{figure*}[t!]
    \centering
    \includegraphics[width=0.9\textwidth]{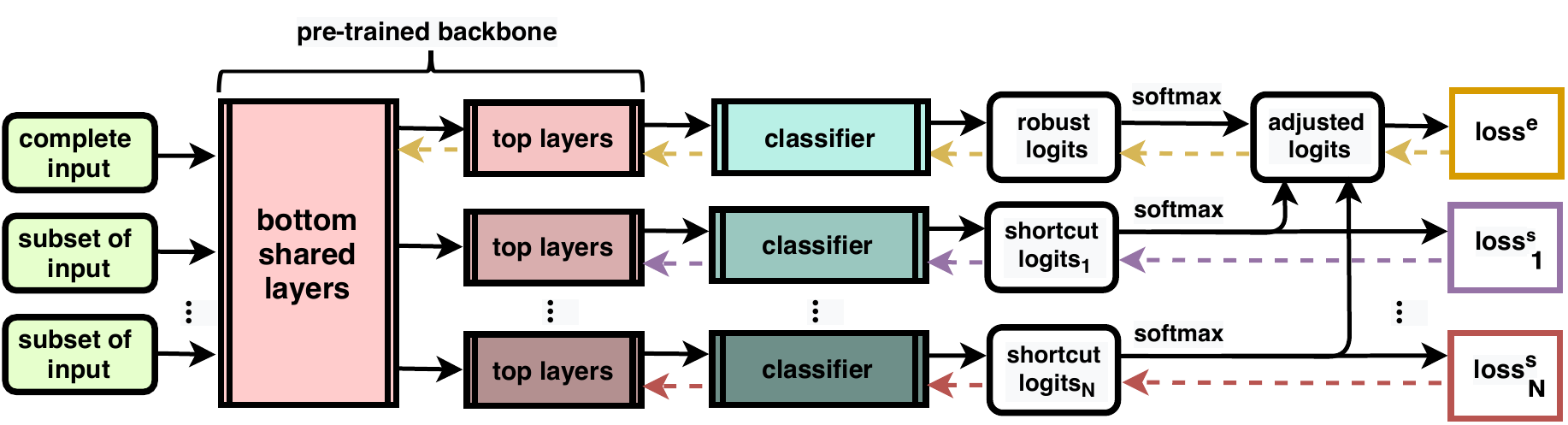}
    \caption{
    Multi-task training framework in our CVC.
    %
    The complete input (\eg, $\mathcal{X}=\{\textit{P}, \textit{Q}, \textit{O}\}$ for MCQA) are fed to the robust branch, while a subset to each shortcut (\eg, $\mathcal{X}_n=\{\textit{P},\textit{O}\}$ to the $n$-th branch). 
    Solid arrows indicate feedforward, and dashed arrows for backpropagation.
    }
    \label{fig_framework}
    \vspace{-3mm}
\end{figure*}


We introduce how to implement CVC in deep neural networks. We have two subsections: one for CVC training and the other for CVC-IV and CVC-MV inferences using trained models. We highlight that CVC training is just the supervised training on multi-task networks which is not novel~\cite{cadene2019rubi, clark2019don}, some of works on other tasks using similar architecture, but our CVC-IV and CVC-MV inference methods are derived from our causal analysis of QA models --- our main contribution to QA methodology.
%

%
%

\subsection{Multi-task Training}
\label{training stage}
%
As illustrated in Figure~\ref{fig_framework}, we deploy the state-of-the-art BERT-based QA model~\cite{devlin2019bert}, as the baseline, and add a few more network branches each of which handles a case of muting part of variables.
We take the main branch having complete variables as input to learn the causal effect corresponding to the robust path of SCM (\ie, $\textit{P},\textit{Q},\textit{O}\! \rightarrow\!\textit{R}\!\rightarrow\!\textit{A}$), so we call it comprehensive reasoning branch (or robust branch).
We use the other branches taking incomplete inputs (part of variables muted) to explicitly learn the shortcut correlations corresponding to the shortcut paths of SCM (\eg, $\textit{P},\textit{O}\!\rightarrow\!\textit{A}$ as $\textit{Q}$ is muted).
We train the model in the manner of multi-task training, \ie, each branch is optimized using an individual objective (the same objective applied across all branches in our case). Only the robust branch gradients can be propagated to update the bottom shared layers in the backbone.
Details are as follows.

\noindent\textbf{Robust branch} is denoted as $F^r$.
It has the complete input denoted as $\mathcal{X}$, \eg, the realistic values of \emph{question}, \emph{passage} and \emph{options} in MCQA.
%
Its network body, with parameters denoted as $\theta^r$, consists of a pre-trained backbone (\eg, BERT) and a classifier (\eg, one FC layer).
Its prediction can be formulated as:
\begin{equation}
    \textit{A}^r = F^r(\mathcal{X}; \theta^r).
    \label{eq_robust_pre}
\end{equation}
In CVC, we need to adjust the prediction $\textit{A}^r$ using shortcut predictions (given by shortcut branches), following the related work~\cite{cadene2019rubi}.
We will elaborate the details and explanations in the paragraph of \textbf{Loss Computation}.

\noindent\textbf{Shortcut branches}
aim to explicitly 
learn the unrobust correlations between incomplete (controlled) input and the ground truth answer. 
%
Each branch is denoted as $F^s_n$ ($n\!=\!1,2,\cdots,N$), and takes a subset of variables (denoted as $\mathcal{X}_n$) as input, setting the other variables as null.
Its network, with parameters denoted as $\theta^s_n$, has the same architecture with the robust branch.
Its prediction can be formulated as:
\begin{equation}
    \textit{A}^s_n = F^s_n(\mathcal{X}_n;\theta^s_n),
    \label{eq_shortcut_pre}
\end{equation}
where $\mathcal{X}_n\subset\mathcal{X}$. $A^s_n$ is used in two ways: (i) computing the individual QA loss to optimize the $n$-th shortcut branch; and (ii) adjusting $A^r$, see following details.

\noindent\textbf{Loss Computation.}
For the $n$-th \textbf{shortcut branch}, we compute its loss as:
\begin{equation}
    \mathcal{L}_n^s=-\sum_i{\mathtt{p}_i\log\softmax(\textit{A}_{n,i}^s)},
\end{equation}
where $i$ denotes the $i$-th dimension of the prediction,
and $\mathtt{p}$ is the ground truth (one-hot).

For the \textbf{robust branch}, using $\textit{A}^r$ to compute the loss cannot guarantee the model to learn only comprehensive reasoning, because backpropagating its loss simply makes the model to learn overall correlations as in the convention QA models.
We solve this problem by adjusting $\textit{A}^r$ using shortcut predictions $\textit{A}_n^s$. In this way, we can force $\textit{A}^r$ to \emph{only preserve the prediction that can never be achieved by shortcuts}, \ie, the comprehensive reasoning prediction with the complete input variables as input.
We implement this ``adjustment'' by using the following function:
\begin{equation}
    \textit{A}_i^e=\sum_n \hat{\mathtt{p}}_{i}^r\cdot \hat{\mathtt{p}}_{n,i}^s,
    \label{p-fusion}
\end{equation}
where $\hat{\mathtt{p}}_i^r\!=\!\softmax(\textit{A}_i^r)$, $\hat{\mathtt{p}}_{n,i}^s\!=\!\softmax(\textit{A}_{n,i}^s)$ and $i$ is the dimension index of prediction logits. Here we use probabilities instead of logits because we found negative values in logits may reverse the adjustment~\cite{clark2019don} and probabilities ensure each item in Eq.~\ref{p-fusion} are of the same scale.
We then use the adjusted result $\textit{A}^{e}$ to compute the cross-entropy loss for the robust branch:
\begin{equation}
    \mathcal{L}^e
    =-\sum_i{\mathtt{p}_i\log\softmax\big(\sum_n{\hat{\mathtt{p}}_{i}^r\cdot \hat{\mathtt{p}}_{n,i}^s}\big)}.
    \label{direct_l}
\end{equation}

We also find that using Eq.~\ref{direct_l} may cause the robust branch to focus on only the learning of hard samples. We resolve this by using two loss variants:
\begin{equation}
\begin{aligned}
    & \mathcal{L}^{e1}=-\sum_n\sum_i{\mathtt{p}_i\log\mathrm{softmax}(\hat{\mathtt{p}}_{i}^r\cdot\hat{\mathtt{p}}_{n,i}^s)}, \\
    & \mathcal{L}^{e2}=-\sum_n{w_n}\sum_i{\mathtt{p}_i\log\mathrm{softmax}(\hat{\mathtt{p}}_{i}^r\cdot\hat{\mathtt{p}}_{n,i}^s)},
    \label{l_fused}
\end{aligned}
\end{equation}
where $w_n\!=\!\mathrm{softmax}(\mathcal{L}_n^s)\!=\! \frac{\exp(\mathcal{L}_n^s)}{\sum_{m=1}^{m=n} \exp(\mathcal{L}_m^s)}$.
$w_n$ is a weight used to explicitly enhance the effect of the $n$-th shortcut branch on the robust branch.
%
Therefore, we formulate the overall loss used in multi-task training as follows,
\begin{equation}
\mathcal{L}^{all}=\mathcal{L}^{e}+\sum_n\mathcal{L}_n^s.
\end{equation}
where $\mathcal{L}^e$ can be replaced with $\mathcal{L}^{e1}$ or $\mathcal{L}^{e2}$.

\subsection{Counterfactual Inferences}
Counterfactual inference is different from the conventional inference as it does not directly uses the model prediction.
In this section, we explain how to conduct CVC-IV and CVC-MV inferences given the trained models: robust model $F^{r}$ as well as shortcut models $\{F^{s}_n\}^{N}_{n=1}$.

Following the notation formats of NP and CP in Eq.~\ref{e1-1} and Eq.~\ref{e1-2} along with the notation of output for each branch in Eq.~\ref{eq_robust_pre} and Eq.~\ref{eq_shortcut_pre},
we can denote
(i) the prediction of the $n$-th shortcut branch 
as
$a^{s}_n\!=\!F^{s}_n(p,o;\theta^s_n)$ and its muted value as $a^{s*}_n\!=\!F^{s}_n(p^*,o^*;\theta^s_n)$; and (ii)
the prediction of the
robust branch as $a^{r}\!=\!F^{r}(p,q,o;\theta^r)$ and
its muted value as
$a^{r*}\!=\!F^{r}(p^*,q^*,o^*;\theta^r)$.

In the CVC-IV inference, we mute all input variables,
so we get NP as $\textit{A}_{a^{s*}_1,\cdots,a_N^{s*},a^{r*}}$ and CP as $\textit{A}_{a^{s*}_1,\cdots,a_N^{s*},a^r}$.
Using these to replace the NP and CP in Eq.~\ref{cvc-iv} and taking Eq.~\ref{p-fusion} into the replaced Eq.~\ref{cvc-iv}, we can derive the \textbf{CVC-IV inference result} as:
\begin{equation} 
\begin{aligned}
\mathrm{CVC\mbox{-}IV}&=A_{a^{s*}_1,\cdots,a_N^{s*},a^r}-A_{a^{s*}_1,\cdots,a_N^{s*},a^{r*}} \\
&=\sum_{n}\hat{\mathtt{p}}^{r}\cdot c_n^s-\sum_{n}c_{n}^r\cdot c_n^s,
\end{aligned}
\label{imp-cvc-iv}
\end{equation}
where each element in $c_{n}^r$ or $c_n^s$ is the same constant in $[0,1]$.
We highlight that CVC-IV inference corresponds to computing Natural Indirect Effect (NIE) in causal inference~\cite{pearl2018book,pearl2001direct}. It is equivalent to the normal inference on the robust model, similar to existing works such as Learned-Mixin~\cite{clark2019don}. While, CVC-IV is totally derived from the systematical causal analysis in QA 
and is thus more explainable than Learned-Mixin which is heuristic.


In the CVC-MV inference, 
we mute $A^r$, denoted as $a^{r*}$.  
We can denote the NP as $\textit{A}_{a^{s}_1,\cdots,a_N^{s},a^{r}}$, and the CP as $\textit{A}_{a^{s}_1,\cdots,a_N^{s},a^{r*}}$.
Using these to replace the NP and CP in Eq.~\ref{cvc-mv} and taking Eq.~\ref{p-fusion} into the replaced Eq.~\ref{cvc-mv}, we can derive the \textbf{CVC-MV inference result} as:
\begin{equation}
\begin{aligned}
    \mathrm{CVC\mbox{-}MV}&=A_{a_1^s,...,a_N^s,a^r}-A_{a_1^s,...,a_N^s,a^{r*}} \\
    &=\sum_{n}\hat{\mathtt{p}}^{r}\cdot \hat{\mathtt{p}}_{n}^s-\sum_{n}c_n^r\cdot \hat{\mathtt{p}}_{n}^s
\end{aligned}
\label{imp-cvc-mv}
\end{equation}
which is an indirect way of making inference using only the robust branch.
It corresponds to computing Controlled Indirect Effect (CIE) in causal inference~\cite{pearl2018book,pearl2001direct}.

From the empirical results of CVC-MV, we find that the hyperparameter $c_n^r$ makes a clear effect.
So we additionally train a $c$-adaptor (denoted as $F_n^c$) to generate $c_n^r$ automatically.
This can be formulated as:
\begin{equation}
    c_{n}^r=F^c_n(\hat{\mathtt{p}}^{r},\hat{\mathtt{p}}_{n}^s,Distance; \theta^c_n),
    \label{c-adp}
\end{equation}
where $F^c_n(x_1,x_2,x_3;\theta^c_n)\!=\!\mathbf{W}_n^2\tanh(\mathbf{W}_n^1[x_1;x_2;x_3])$, $[;]$ is the concatenation operation, and $\theta^c_n\!=\!\{\mathbf{W}_n^1,\mathbf{W}_n^2\}$ are the parameters of $c$-adaptor. $Distance=\mathbf{JS}[\hat{\mathtt{p}}^{r}||\hat{\mathtt{p}}_{n}^s]$,
where $\mathbf{JS}$ is Jensen-Shannon divergence~\cite{lin1991divergence}. In experiments, we implement $c$-adaptor using a two-layer MLP and conduct an ablative study to show its efficiency.

\section{Experiments}
\label{experiment}

\begin{table}[h]
\centering
\footnotesize
\setlength\tabcolsep{2.1pt}
\begin{tabular}{lllll}
\toprule
                & \textbf{MCTest}          & \textbf{DREAM}     & \textbf{RACE}         & \textbf{SQuAD}     \\\midrule
Construction    & Crowd.          & Exams     & Exams        & Crowd.    \\
Passage type    & Child's stories & Dialogues & Written text & Wikipedia \\
\# of passages  & 660             & 6,444     & 27,933       & 23,215    \\
\# of questions & 2,640           & 10,197    & 97,687       & 107,785   \\
\# of options   & 4               & 3         & 4            & -  \\\bottomrule       
\end{tabular}
\caption{
We conduct MCQA experiments on three datasets, \ie, MCTest~\cite{richardson2013mctest}, DREAM~\cite{sun2019dream}, RACE~\cite{lai2017race}, and SEQA experiments on the SQuAD dataset~\cite{rajpurkar2016squad}.
``Crowd.'': crowd-sourcing; ``-'': not applicable.
}
\label{dataset_statistic}
\end{table}
We evaluate the robustness of CVC for both MCQA and SEQA, using a variety of adversarial attacks~\cite{zhang2020adversarial}.
Specifically, MCQA aims to predict the correct answer from several input options given a passage and a question.
SEQA locates the answer span in a passage given a question.
We show the dataset information in Table~\ref{dataset_statistic}.
Below we introduce adversarial attacks, implementation details, an ablation study, a case study and the comparison to state-of-the-arts.
\begin{table*}[t]
\footnotesize
\centering
\setlength\tabcolsep{4pt}
\begin{tabular}{llllllllllllll}
\toprule 
\multicolumn{1}{l}{}                        & \multicolumn{1}{l}{}         & \multicolumn{6}{c}{\textbf{BERT-base}}                                                                                                            & \multicolumn{6}{c}{\textbf{BERT-large}}                                                                                                           \\ 
                       \cmidrule(r){3-8}
                       \cmidrule(r){9-14}
Dataset & Method    & \texttt{Test}                      & \texttt{Adv1}                      & \texttt{Adv2}                      & \texttt{Adv3}                      & \texttt{Adv4}                      & A.G.                      & \texttt{Test}                      & \texttt{Adv1}                      & \texttt{Adv2}                      & \texttt{Adv3}                      & \texttt{Adv4}                      &A.G.            \\ \midrule
\multicolumn{1}{l}{\multirow{3}{*}{\textbf{MCTest}}} & \multicolumn{1}{l}{CT~\cite{devlin2019bert}} & \multicolumn{1}{l}{68.9} & \multicolumn{1}{l}{63.9} & \multicolumn{1}{l}{59.4} & \multicolumn{1}{l}{20.2} & \multicolumn{1}{l}{54.8} & \multicolumn{1}{l}{-} & \multicolumn{1}{l}{72.3} & \multicolumn{1}{l}{70.0}   & \multicolumn{1}{l}{66.8} & \multicolumn{1}{l}{35.5} & \multicolumn{1}{l}{57.6} & \multicolumn{1}{l}{-} \\ 
\multicolumn{1}{c}{}                        & CVC-MV                      & 68.1                     & 69.1                      & \textbf{65.6}                     & 26.8                      & \textbf{61.0}         & +6.1\%               & 73.2                          & 74.3                       & 73.5                          & 38.4                          & 68.4             & +6.2\%              \\
\multicolumn{1}{c}{}                        & CVC-IV                           & \textbf{69.4}                      & \textbf{70.0}                        & 65.4                      & \textbf{28.7}                      & 59.9        & \textbf{+6.4\%}              & \textbf{74.4}                      & \textbf{75.5}                      & \textbf{75.1}                      & \textbf{40.4}                      & \textbf{69.5}             & \textbf{+7.6\%}         \\
 \midrule
\multicolumn{1}{l}{\multirow{3}{*}{\textbf{DREAM}}}  & \multicolumn{1}{l}{CT~\cite{devlin2019bert}} & \multicolumn{1}{l}{\textbf{61.5}} & \multicolumn{1}{l}{47.5} & \multicolumn{1}{l}{39.2} & \multicolumn{1}{l}{20.9} & \multicolumn{1}{l}{41.8} & \multicolumn{1}{l}{-} & \multicolumn{1}{l}{\textbf{65.9}} & \multicolumn{1}{l}{50.6} & \multicolumn{1}{l}{43.0}   & \multicolumn{1}{l}{25.6} & \multicolumn{1}{l}{48.2} & \multicolumn{1}{l}{-} \\  
\multicolumn{1}{c}{}                        & CVC-MV                      & 60.1                      & \textbf{49.6}                      & 39.9                      & 23.7                      & 45.6        & +2.3\%              & 64.0                          & 51.9                          & \textbf{46.5}                          & 26.3                          & \textbf{51.3}        & \textbf{+2.2\%} \\
& CVC-IV                           & 60.0                        & 49.2                      & \textbf{40.7}                      & \textbf{25.0}                        & \textbf{47.1}       & \textbf{+3.1\%}                   & 64.5                      & \textbf{52.0}                        & 46.2                      & \textbf{26.6}                      & 51.1          & +2.1\%            \\
                \midrule
\multicolumn{1}{l}{\multirow{3}{*}{\textbf{RACE}}}   & \multicolumn{1}{l}{CT~\cite{devlin2019bert}} & \multicolumn{1}{l}{\textbf{64.7}} & \multicolumn{1}{l}{56.0}   & \multicolumn{1}{l}{50.1} & \multicolumn{1}{l}{36.6} & \multicolumn{1}{l}{58.3} & \multicolumn{1}{l}{-} & \multicolumn{1}{l}{67.9} & \multicolumn{1}{l}{61.9} & \multicolumn{1}{l}{57.9} & \multicolumn{1}{l}{51.0}   & \multicolumn{1}{l}{61.7} & \multicolumn{1}{l}{-} \\ 
\multicolumn{1}{c}{}                        & CVC-MV                      & 64.4                          & 56.7                          & 51.7                          & \textbf{39.1}                          & \textbf{59.2}         & \textbf{+1.4\%}                 & \textbf{68.5}                          & 62.6                          & 58.2                           & \textbf{52.0}                           & \textbf{65.7}          & +1.5\%                \\
& CVC-IV                           & 64.1                      & \textbf{57.0}                        & \textbf{52.2}                      & 38.8                      & 58.6   & +1.4\%                    & 68.4                          & \textbf{63.1}                          & \textbf{59.1}                         & 51.3                          & 65.1     & \textbf{+1.6\%}                     \\
 \bottomrule 
\end{tabular}
\caption{
Accuracies (\%) on three MCQA datasets.
Models are trained on original training data.
BERT-base and BERT-large are backbones.
``A.G.'' denotes the average improvement over the conventional training (CT)~\cite{devlin2019bert} for \texttt{Adv*} sets.}
\label{table:main_results_MCQA}
\end{table*}
\begin{table*}[t]
\footnotesize
\centering
\begin{tabular}{lllllllllll}
\toprule           & \multicolumn{5}{c}{\textbf{BERT-base}} & \multicolumn{5}{c}{\textbf{BERT-large}} \\\cmidrule(r){2-6} \cmidrule(r){7-11}
 Method  & \texttt{Dev}   & \texttt{Adv1}  & \texttt{Adv2}  & \texttt{Adv3}  & A.G. & \texttt{Dev}    & \texttt{Adv1}  & \texttt{Adv2}  & \texttt{Adv3} & A.G.  \\
\midrule 
CT~\cite{devlin2019bert} (by~\cite{liurobust})         & 88.4  & 49.9  & 59.7$^*$  & 44.6$^*$ & - & \textbf{90.6}   & 60.2  & 70.0$^*$  & 50.0$^*$ & -\\
QAInformax~\cite{yeh2019qainfomax} & \textbf{88.6}  & 54.5  & 64.9  & -   & +4.9\%  & -      & -     & -     & -  & -   \\
CVC-MV   & 87.2  & 55.7  & 65.3  & 51.3 & +6.0\% & 90.2   & 62.6  & \textbf{72.4}  & 52.5 & +2.4\%\\
CVC-IV      & 86.6  & \textbf{56.3}  & \textbf{66.2}  & \textbf{51.5} & \textbf{+6.6\%} & 89.4   & \textbf{62.6}  & 71.8  & \textbf{54.1} & \textbf{+2.8\%} \\
\bottomrule 
\end{tabular}
\caption{SEQA F1-measure (\%) on the SQuAD \texttt{Dev} set (\texttt{Test} set is not public) and adversarial sets.
Models are trained on original training data.
BERT-base and BERT-large are backbones.
``-'': not applicable.
``*'': our implementation using the public code.
``A.G.'': our average improvement over the conventional training (CT)~\cite{devlin2019bert} for \texttt{Adv*}.}
\label{table:main_results_SEQA}
\end{table*}

\noindent\textbf{Adversarial Attacks on MCQA.} 
We propose $4$ kinds of grammatical adversarial attacks to generate $4$ sets of adversarial examples (using original test sets as bases), respectively. 
\texttt{Add1Truth2Opt} (\texttt{Adv1}) (and \texttt{Add2Truth2Opt} (\texttt{Adv2})): 
we replace one (and two) of the wrong options with one (and two) of the correct ones borrowed from other samples attached to the same passage.
\texttt{Add1Pas2Opt} (\texttt{Adv3}): 
We replace one of the wrong options with a random distractor sentence extracted from the passage. 
Note that this distractor does not overlap with the ground truth option except stop words or punctuation marks.
\texttt{Add1Ent2Pas} (\texttt{Adv4}): 
We first choose one of the wrong options with at least one entity, and then replace each entity with another entity of the same type.
Then, we add this modified option (as a sentence) to the end of the passage.

\noindent\textbf{Adversarial Attacks on SEQA.\dag} 
We utilize $3$ kinds of grammatical adversarial attacks
(using the original \texttt{Dev} set as basis), denoted as
\texttt{AddSent} (\texttt{Adv1}), \texttt{AddOneSent} (\texttt{Adv2}) and \texttt{AddVerb} (\texttt{Adv3}). 
\texttt{AddSent} and \texttt{AddOneSent} released by~\cite{jia2017adversarial} add distracting sentences 
to the passage.
They can be used to measure the model robustness against entity or noun attacks.
In addition, we propose \texttt{AddVerb} for which we hire an expert linguist to annotate the data.
Examples are as follows. For the question \textit{``What city did Tesla move to in 1880?''}, 
\texttt{AddSent} sample could be \textit{``Tadakatsu moved to the city of Chicago in 1881.''}, and 
\texttt{AddVerb} sample could be \textit{``Tesla left the city of Chicago in 1880.''}.
We use \texttt{AddVerb} to
evaluate 
the model robustness against verb attacks.

\noindent\textbf{Implementation Details.\dag}
We deploy the pre-trained BERT backbones provided by HuggingFace~\cite{Wolf2019HuggingFacesTS}.
Following~\cite{clark2019don,grand2019adversarial,ramakrishnan2018overcoming},
we perform model selection (\ie, choosing the hyperparameters of training epochs) based on the model performance in the development/test sets on the used dataset. 

\noindent\textbf{MCQA-Specific.\dag} 
MCQA has two shortcut correlations (see Figure~\ref{fig_human_machine}), \ie, $\textit{Q}\!\rightarrow\!\textit{A}$ and $\textit{P}\!\rightarrow\!\textit{A}$ ($\textit{O}\!\rightarrow\!\textit{A}$
is not discussed here as \textit{O} is mandatory and can not be muted).
We inspect them and notice that the effect from the former one is trivial and negligible compared to the latter. 
One may argue that $\textit{Q}$ is an important cue to predict the answer.
While the fact is when building MCQA datasets, annotators intentively avoid any easy question-answer data.
For example, 
they include a person name in all options of the \textit{who} question.
We thus assume $\textit{Q}\!\rightarrow\!\textit{A}$ has been eliminated during well-designed data collection and utilize one shortcut branch (muting \textit{Q}).
Eq.~\ref{direct_l} and Eq.~\ref{l_fused} are equivalent for
MCQA ($N\!=\!1$ and $w_n\!=\!1$).
%
%

\noindent\textbf{SEQA-Specific.\dag} 
Different from MCQA, we propose to manually separate the \emph{question} (\textit{Q}) of SEQA into corresponding parts:
entities \& nouns (\textit{E)}; verbs \& adverbs (\textit{V}); and the remaining stop words
\& punctuation marks (\textit{S}).
%
Therefore, the SCM of SEQA contains four input variables as \textit{P} (\emph{passage}), \textit{E}, \textit{V} and \textit{S}.
The comprehensive \emph{reasoning} variable \textit{R} mediates between these four variables and \emph{answer} \textit{A}.
We inspect the empirical effects of all shortcut paths, and build shortcut branches with $N\!=\!2$:
one branch with \textit{E} muted, and the other with \textit{V} muted. We use $\mathcal{L}^{e2}$ to train SEQA models.

\subsection{Results and Analyses}

\begin{figure}
   \centering
    \includegraphics[width=0.47\textwidth]{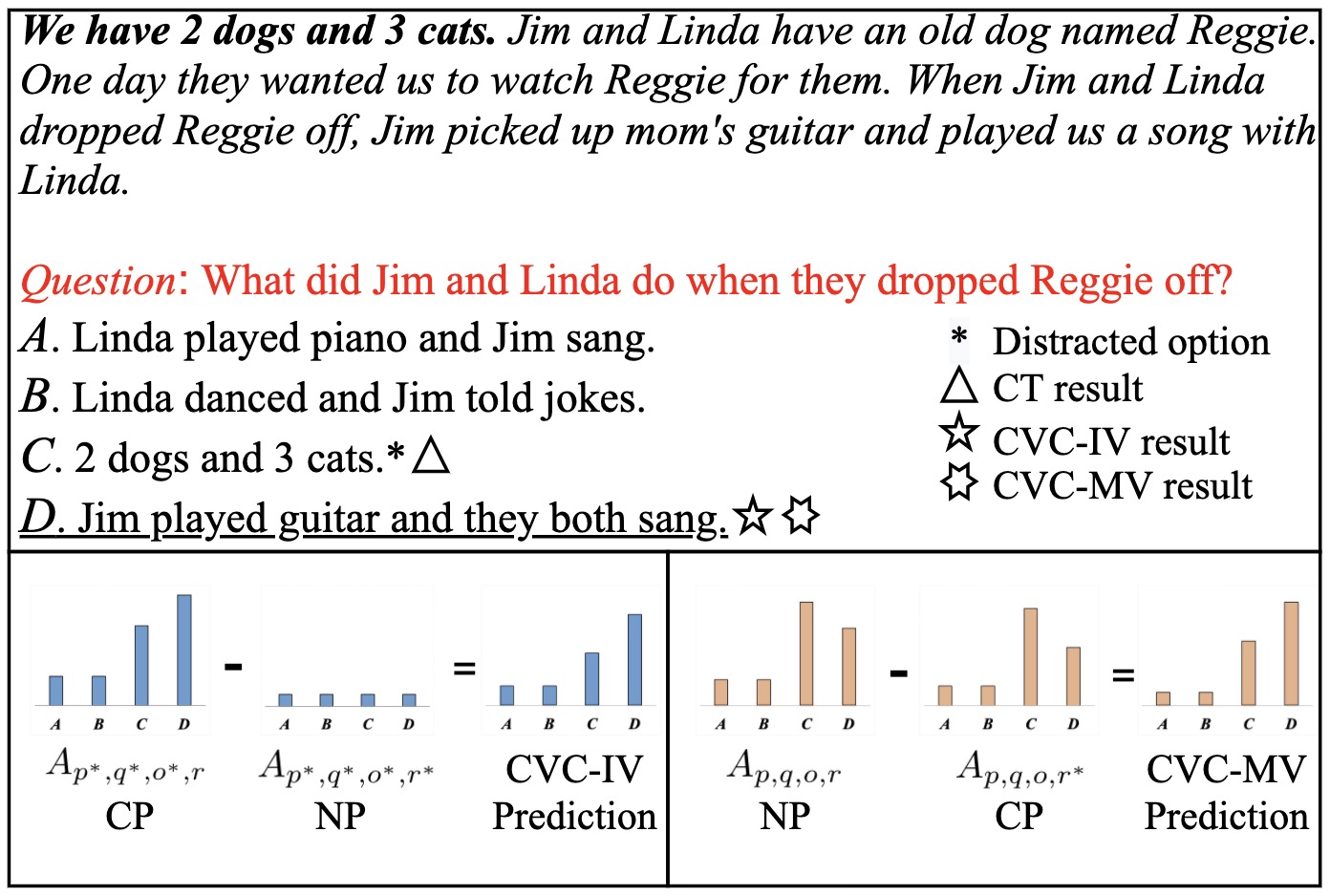}
    \caption{A case study of CVC on MCTest trained on official data. The ground truth is \underline{underlined}.}
    \label{case_study}
    \vspace{-2mm}
\end{figure}

Table~\ref{table:main_results_MCQA} and Table~\ref{table:main_results_SEQA} show the overall results for MCQA and SEQA, respectively.
Table~\ref{table:main_results_augment_} particularly compares ours to ensembling based methods.
Table~\ref{ablation} presents an ablation study on SEQA.
Figure~\ref{case_study} shows an case study of MCQA.

\begin{table}[h]
\centering
\footnotesize
\begin{tabular}{lllll}
\toprule
              & MCTest & DREAM & RACE & SQuAD \\\midrule
DRiFt         & 1.9    & 2.5   & \textbf{1.7}  & 4.5    \\
Bias Prodcut  & 5.1    & -0.1  & 1    & 4.1$^*$       \\
Learned-Mixin & 1.8    & 1.0     & 1.4  & 2.1$^*$      \\ \midrule
CVC-MV(ours)        & 6.1    & 2.3   & 1.4  & 6.0    \\
CVC-IV(ours)        & \textbf{6.4}    & \textbf{3.1}   & 1.4  & \textbf{6.6}   \\
\bottomrule
\end{tabular}
\caption{
Comparing to related ensemble-based methods: DRiFT~\cite{he2019unlearn}, Bias Product and learned-Mixin~\cite{clark2019don} with BERT-base. (\emph{We show A.G. only, due to the page limits}).
We implement related methods by replacing Eq.~\ref{p-fusion} with their adjustment functions. ``*'': the design of shortcut branch in original paper is used (one shortcut branch on SQuAD), otherwise our design is used (two shortcut branches on SQuAD).
}
\label{table:main_results_augment_}
\end{table}
\noindent\textbf{Overall Results Compared to Baselines and State-of-the-Art.\dag}
From Table~\ref{table:main_results_MCQA}, we can see that both CVC-MV and CVC-IV can surpass the baseline method~\cite{devlin2019bert} for defending against adversarial attacks, \eg, by average accuracies of $7\%$ and $1.5\%$ on MCTest and RACE, respectively.
It is worth highlighting the example that CVC-IV on BERT-base gains $8.5\%$ on the most challenging \texttt{Adv3} set of MCTest.
These observations are consistent in the results of SEQA, shown in Table~\ref{table:main_results_SEQA}.
Besides, it is clear that ours outperforms the state-of-the-art QAInformax~\cite{yeh2019qainfomax}, \eg, by an average of $1.7\%$ F1-measure (on the same BERT-base backbone). Ours also outperforms model ensembling based methods in most of the datasets, as shown in Table~\ref{table:main_results_augment_}.
Please note that these methods can be regarded as implementation-level examples under our CVC-IV formulation.
%
%
From Table~\ref{table:main_results_MCQA}-\ref{table:main_results_augment_}, we also notice that 
CVC-MV often performs worse than CVC-IV on \texttt{Adv*} sets but better on in-domain \texttt{Test} (or \texttt{Dev}) sets.
The possible reason is that the important hyperparameter of CVC-MV $c_n^r$ is learned from in-domain data.
Please refer to our supplementary materials where we show that augmenting in-domain data with \texttt{Adv*} examples greatly improves the performance of CVC-MV.

\begin{table}[h]
\centering
\footnotesize
\begin{tabular}{lllll}
\toprule 
Ablative Setting & \texttt{Dev}  & \texttt{Adv1} & \texttt{Adv2} & \texttt{Adv3} \\\midrule
(1) w/o first Shct.br. & 85.5 & 52.6 & 62.5 & 50.8 \\
(2) w/o second Shct.br. & 86.1 & \textbf{57.7} & 66.1 & 42.1 \\
(3) use $\mathcal{L}^{e}$          & 72.4 & 45.9 & 54.9 & 42.6 \\
(4) use $\mathcal{L}^{e1}$         & 86.5 & 53.5 & 63.2 & 46.7 \\ 
\midrule
CVC-IV (ours)        & \textbf{86.6} & 56.3 & \textbf{66.2} & \textbf{51.5} \\ 
\midrule[0.8pt]
(5) same $c_{r,n}$   & 85.7 & 54.3 & 64.1 & 51.0 \\ 
(6) $c_{r,n}\!=\!JS$ & 85.9 & 54.3 & 64.2 & 51.1 \\
(7) $c_{r,n}\!=\!Euc$ & 86.0 & 54.4 & 64.1 & 51.2 \\
(8) w/o $distance$  & 86.9 & 55.3 & 65.0 & 51.3 \\
(9) w/o $\hat{p}_{r}$ and $\hat{p}_{n}$ & 84.0 & 53.2 & 62.6 & 49.4 \\
(10) features as input & \textbf{87.6} & 55.0 & 64.9  & 50.9 \\
\midrule
CVC-MV (ours)    & 87.2 & \textbf{55.7} & \textbf{65.3} & \textbf{51.3} \\ \bottomrule 
\end{tabular}
\caption{The ablation study on SQuAD (BERT-base). (1)-(3) are ablative settings for multi-task training (using CVC-IV);
(4)-(9) are ablative settings related to CVC-MV.}
\label{ablation}
\end{table}
\noindent\textbf{Ablation Study.\dag}
In Table~\ref{ablation}, we show the SEQA results in $10$ ablative settings, to evaluate the approach when:
(1) remove the first shortcut branch ($\textit{E}$ muted) from the multi-task training;
(2) remove the second shortcut branch ($\textit{V}$ muted) from the multi-task training; (3) use $\mathcal{L}^e$ to replace $\mathcal{L}^{e2}$; (4) use $\mathcal{L}^{e1}$ to replace $\mathcal{L}^{e2}$;
(5) set $c_n^r$ to the same constant (tuned in $\{0.2,0.4,0.6,0.8,1\}$) for all input samples;
(6) let $c_n^r=\mathbf{JS}[\hat{p}^{r}||\hat{p}_{n}^s]$;
(7) let $c_n^r=\left|\hat{p}^{r}-\hat{p}_{n}^s\right|^2/2$ where $2$ is the upper bound of the square of Euclidean distance;
(8) remove $distance$ item in Eq.\ref{c-adp};
(9) remove $\hat{p}^{r}$ and $\hat{p}_{n}^s$ in Eq.~\ref{c-adp};
and (10) use the features extracted via top layers to replace $\hat{p}^{r}$, $\hat{p}_{n}^s$ and $\mathbf{JS}$ distance in Eq.~\ref{c-adp}.




Compared to the ablative results, we can see that our full approach achieves the overall top performance.
Two exceptions are (i) a higher score is achieved for \texttt{Adv1} if without the second shortcut branch and (ii) a slight higher score for \texttt{Dev} if replacing $\hat{p}^{r}$ and $\hat{p}_{n}^s$ with features. 
We think (i) is because the model missing a shortcut branch simply over-fits to the corresponding shortcut correlation (bias).
For (ii), a possible reason is that  higher-dimensional features (of in-domain data) have stronger representation ability than probability vectors.

\noindent\textbf{Case Study.\dag}
We visualize an example of MCQA in
Figure~\ref{case_study} to demonstrate the underlying mechanism of CVC-IV and CVC-MV inference.
CT~\cite{devlin2019bert} merely aligns the words between \emph{passage} and \emph{options},
which leads to the wrong choice $C$ (a confusing choice generated by \texttt{Adv1}).
In contrast,
both our CVC-IV and CVC-MV pick the right answer $D$.
On the bottom blocks, we 
demonstrate the calculation on prediction logits during CVC-IV (Eq.~\ref{cvc-iv}) and CVC-MV (Eq.~\ref{cvc-mv}), respectively.
We take the CVC-MV as an example to interpret this calculation.
Both Normal Prediction (NP) $\textit{A}_{p,q,o,r}$ and Counterfactual Prediction (CP) $\textit{A}_{p,q,o,r^*}$ contain the logits of $A$, $B$, $C$ and $D$.
The logit value of $C$ is from the word alignment shortcut and it is high in both NP and CP.
It thus can be counteracted after the subtraction in CVC-MV.
In contrast, the logit value of $D$ is from the comprehensive \emph{reasoning}. When muting the corresponding variable $\textit{R}$ (denoted by $r^*$ in CP 
$\textit{A}_{p,q,o,r^*}$), this value must be reduced.
Then it becomes evident after the subtraction in CVC-MV. Please note that we normalize the bar chart (the result of the subtraction) to provide a clear visualization.

\section{Conclusions}
We inspect the problem of fragility in QA models, and build the structural causal model to show that the crux is from shortcut correlations.
To train robust QA models, we propose a novel CVC approach and implement it on the multi-task training pipeline.
We conduct extensive experiments on a variety of QA benchmarks, and show that our approach can achieve high robustness and good interpretation.
Our future work is to enhance the structural causal model by considering the subjective factors, \eg, the preference of dataset annotators and the source of passages.

\bibliography{reference}

\begin{thebibliography}{62}
\providecommand{\natexlab}[1]{#1}
\providecommand{\url}[1]{\texttt{#1}}
\providecommand{\urlprefix}{URL }
\expandafter\ifx\csname urlstyle\endcsname\relax
  \providecommand{\doi}[1]{doi:\discretionary{}{}{}#1}\else
  \providecommand{\doi}{doi:\discretionary{}{}{}\begingroup
  \urlstyle{rm}\Url}\fi

\bibitem[{Abbasnejad et~al.(2020)Abbasnejad, Teney, Parvaneh, Shi, and
  Hengel}]{abbasnejad2020counterfactual}
Abbasnejad, E.; Teney, D.; Parvaneh, A.; Shi, J.; and Hengel, A. v.~d. 2020.
\newblock Counterfactual vision and language learning.
\newblock In \emph{CVPR}, 10044--10054.

\bibitem[{Alzantot et~al.(2018)Alzantot, Sharma, Elgohary, Ho, Srivastava, and
  Chang}]{alzantot2018generating}
Alzantot, M.; Sharma, Y.; Elgohary, A.; Ho, B.-J.; Srivastava, M.; and Chang,
  K.-W. 2018.
\newblock Generating Natural Language Adversarial Examples.
\newblock In \emph{EMNLP}, 2890--2896.

\bibitem[{Bowman et~al.(2015)Bowman, Angeli, Potts, and
  Manning}]{bowman2015large}
Bowman, S.; Angeli, G.; Potts, C.; and Manning, C.~D. 2015.
\newblock A large annotated corpus for learning natural language inference.
\newblock In \emph{EMNLP}, 632--642.

\bibitem[{Bras et~al.(2020)Bras, Swayamdipta, Bhagavatula, Zellers, Peters,
  Sabharwal, and Choi}]{bras2020adversarial}
Bras, R.~L.; Swayamdipta, S.; Bhagavatula, C.; Zellers, R.; Peters, M.~E.;
  Sabharwal, A.; and Choi, Y. 2020.
\newblock Adversarial Filters of Dataset Biases.
\newblock \emph{arXiv preprint arXiv:2002.04108} .

\bibitem[{Cadene et~al.(2019)Cadene, Dancette, Cord, Parikh
  et~al.}]{cadene2019rubi}
Cadene, R.; Dancette, C.; Cord, M.; Parikh, D.; et~al. 2019.
\newblock RUBi: Reducing Unimodal Biases for Visual Question Answering.
\newblock In \emph{NeurIPS}, 839--850.

\bibitem[{Chen et~al.(2020)Chen, Yan, Xiao, Zhang, Pu, and
  Zhuang}]{chen2020counterfactual}
Chen, L.; Yan, X.; Xiao, J.; Zhang, H.; Pu, S.; and Zhuang, Y. 2020.
\newblock Counterfactual Samples Synthesizing for Robust Visual Question
  Answering.
\newblock \emph{arXiv preprint arXiv:2003.06576} .

\bibitem[{Clark, Yatskar, and Zettlemoyer(2019)}]{clark2019don}
Clark, C.; Yatskar, M.; and Zettlemoyer, L. 2019.
\newblock Don’t Take the Easy Way Out: Ensemble Based Methods for Avoiding
  Known Dataset Biases.
\newblock In \emph{EMNLP}, 4060--4073.

\bibitem[{Devlin et~al.(2019)Devlin, Chang, Lee, and
  Toutanova}]{devlin2019bert}
Devlin, J.; Chang, M.-W.; Lee, K.; and Toutanova, K. 2019.
\newblock BERT: Pre-training of Deep Bidirectional Transformers for Language
  Understanding.
\newblock In \emph{NAACL}, 4171--4186.

\bibitem[{Ebrahimi, Lowd, and Dou(2018)}]{ebrahimi2018adversarial}
Ebrahimi, J.; Lowd, D.; and Dou, D. 2018.
\newblock On Adversarial Examples for Character-Level Neural Machine
  Translation.
\newblock In \emph{COLING}, 653--663.

\bibitem[{Ebrahimi et~al.(2018)Ebrahimi, Rao, Lowd, and
  Dou}]{ebrahimi2018hotflip}
Ebrahimi, J.; Rao, A.; Lowd, D.; and Dou, D. 2018.
\newblock HotFlip: White-Box Adversarial Examples for Text Classification.
\newblock In \emph{ACL}, 31--36.

\bibitem[{Feng et~al.(2018)Feng, Wallace, Grissom~II, Iyyer, Rodriguez, and
  Boyd-Graber}]{feng2018pathologies}
Feng, S.; Wallace, E.; Grissom~II, A.; Iyyer, M.; Rodriguez, P.; and
  Boyd-Graber, J. 2018.
\newblock Pathologies of Neural Models Make Interpretations Difficult.
\newblock In \emph{EMNLP}, 3719--3728.

\bibitem[{Goyal et~al.(2019)Goyal, Wu, Ernst, Batra, Parikh, and
  Lee}]{goyal2019counterfactual}
Goyal, Y.; Wu, Z.; Ernst, J.; Batra, D.; Parikh, D.; and Lee, S. 2019.
\newblock Counterfactual Visual Explanations.
\newblock In \emph{ICML}, 2376--2384.

\bibitem[{Grand and Belinkov(2019)}]{grand2019adversarial}
Grand, G.; and Belinkov, Y. 2019.
\newblock Adversarial Regularization for Visual Question Answering: Strengths,
  Shortcomings, and Side Effects.
\newblock In \emph{Proceedings of the Second Workshop on Shortcomings in Vision
  and Language}, 1--13.

\bibitem[{He, Zha, and Wang(2019)}]{he2019unlearn}
He, H.; Zha, S.; and Wang, H. 2019.
\newblock Unlearn Dataset Bias in Natural Language Inference by Fitting the
  Residual.
\newblock In \emph{Proceedings of the 2nd Workshop on Deep Learning Approaches
  for Low-Resource NLP (DeepLo 2019)}, 132--142.

\bibitem[{Huang et~al.(2019)Huang, Stanforth, Welbl, Dyer, Yogatama, Gowal,
  Dvijotham, and Kohli}]{huang2019achieving}
Huang, P.-S.; Stanforth, R.; Welbl, J.; Dyer, C.; Yogatama, D.; Gowal, S.;
  Dvijotham, K.; and Kohli, P. 2019.
\newblock Achieving Verified Robustness to Symbol Substitutions via Interval
  Bound Propagation.
\newblock In \emph{EMNLP}, 4074--4084.

\bibitem[{Iyyer et~al.(2018)Iyyer, Wieting, Gimpel, and
  Zettlemoyer}]{iyyer2018adversarial}
Iyyer, M.; Wieting, J.; Gimpel, K.; and Zettlemoyer, L. 2018.
\newblock Adversarial Example Generation with Syntactically Controlled
  Paraphrase Networks.
\newblock In \emph{NAACL}.

\bibitem[{Jia and Liang(2017)}]{jia2017adversarial}
Jia, R.; and Liang, P. 2017.
\newblock Adversarial Examples for Evaluating Reading Comprehension Systems.
\newblock In \emph{EMNLP}, 2021--2031.

\bibitem[{Jia et~al.(2019)Jia, Raghunathan, G{\"o}ksel, and
  Liang}]{jia2019certified}
Jia, R.; Raghunathan, A.; G{\"o}ksel, K.; and Liang, P. 2019.
\newblock Certified Robustness to Adversarial Word Substitutions.
\newblock In \emph{EMNLP}, 4120--4133.

\bibitem[{Kaushik, Hovy, and Lipton(2020)}]{kaushik2020}
Kaushik, D.; Hovy, E.; and Lipton, Z.~C. 2020.
\newblock Learning the Difference that Makes a Difference with
  Counterfactually-Augmented Data.
\newblock In \emph{ICLR}.

\bibitem[{Kaushik and Lipton(2018)}]{kaushik2018much}
Kaushik, D.; and Lipton, Z.~C. 2018.
\newblock How Much Reading Does Reading Comprehension Require? A Critical
  Investigation of Popular Benchmarks.
\newblock In \emph{EMNLP}, 5010--5015.

\bibitem[{Lai et~al.(2017)Lai, Xie, Liu, Yang, and Hovy}]{lai2017race}
Lai, G.; Xie, Q.; Liu, H.; Yang, Y.; and Hovy, E. 2017.
\newblock RACE: Large-scale ReAding Comprehension Dataset From Examinations.
\newblock In \emph{EMNLP}, 785--794.

\bibitem[{Lawrence and Riezler(2018)}]{lawrence2018improving}
Lawrence, C.; and Riezler, S. 2018.
\newblock Improving a Neural Semantic Parser by Counterfactual Learning from
  Human Bandit Feedback.
\newblock In \emph{ACL}, 1820--1830.

\bibitem[{Lin(1991)}]{lin1991divergence}
Lin, J. 1991.
\newblock Divergence measures based on the Shannon entropy.
\newblock \emph{IEEE Transactions on Information theory} 37(1): 145--151.

\bibitem[{Liu et~al.(2020)Liu, Liu, Yang, Liu, Su, Li, and She}]{liurobust}
Liu, K.; Liu, X.; Yang, A.; Liu, J.; Su, J.; Li, S.; and She, Q. 2020.
\newblock A Robust Adversarial Training Approach to Machine Reading
  Comprehension.
\newblock In \emph{AAAI}.

\bibitem[{Liu et~al.(2019)Liu, Ott, Goyal, Du, Joshi, Chen, Levy, Lewis,
  Zettlemoyer, and Stoyanov}]{liu2019roberta}
Liu, Y.; Ott, M.; Goyal, N.; Du, J.; Joshi, M.; Chen, D.; Levy, O.; Lewis, M.;
  Zettlemoyer, L.; and Stoyanov, V. 2019.
\newblock Roberta: A robustly optimized bert pretraining approach.
\newblock \emph{arXiv preprint arXiv:1907.11692} .

\bibitem[{McCoy, Pavlick, and Linzen(2019)}]{mccoy2019right}
McCoy, T.; Pavlick, E.; and Linzen, T. 2019.
\newblock Right for the Wrong Reasons: Diagnosing Syntactic Heuristics in
  Natural Language Inference.
\newblock In \emph{Proceedings of the 57th Annual Meeting of the Association
  for Computational Linguistics}, 3428--3448.

\bibitem[{Morgan and Winship(2015)}]{morgan2015counterfactuals}
Morgan, S.~L.; and Winship, C. 2015.
\newblock \emph{Counterfactuals and causal inference}.
\newblock Cambridge University Press.

\bibitem[{Neuberg(2003)}]{neuberg2003causality}
Neuberg, L.~G. 2003.
\newblock Causality: models, reasoning, and inference.
\newblock \emph{Econometric Theory} 19(4): 675--685.

\bibitem[{Niu et~al.(2020)Niu, Tang, Zhang, Lu, Hua, and
  Wen}]{niu2020counterfactual}
Niu, Y.; Tang, K.; Zhang, H.; Lu, Z.; Hua, X.-S.; and Wen, J.-R. 2020.
\newblock Counterfactual VQA: A Cause-Effect Look at Language Bias.
\newblock \emph{arXiv preprint arXiv:2006.04315} .

\bibitem[{Pearl(2001)}]{pearl2001direct}
Pearl, J. 2001.
\newblock Direct and indirect effects.
\newblock In \emph{Proceedings of the Seventeenth conference on Uncertainty in
  artificial intelligence}, 411--420.

\bibitem[{Pearl and Mackenzie(2018)}]{pearl2018book}
Pearl, J.; and Mackenzie, D. 2018.
\newblock \emph{The book of why: the new science of cause and effect}.
\newblock Basic Books.

\bibitem[{Pearl et~al.(2009)}]{pearl2009causal}
Pearl, J.; et~al. 2009.
\newblock Causal inference in statistics: An overview.
\newblock \emph{Statistics surveys} 3: 96--146.

\bibitem[{Qi et~al.(2019)Qi, Niu, Huang, and Zhang}]{qi2019two}
Qi, J.; Niu, Y.; Huang, J.; and Zhang, H. 2019.
\newblock Two Causal Principles for Improving Visual Dialog.
\newblock \emph{arXiv preprint arXiv:1911.10496} .

\bibitem[{Rajpurkar et~al.(2016)Rajpurkar, Zhang, Lopyrev, and
  Liang}]{rajpurkar2016squad}
Rajpurkar, P.; Zhang, J.; Lopyrev, K.; and Liang, P. 2016.
\newblock SQuAD: 100,000+ Questions for Machine Comprehension of Text.
\newblock In \emph{EMNLP}, 2383--2392.

\bibitem[{Ramakrishnan, Agrawal, and Lee(2018)}]{ramakrishnan2018overcoming}
Ramakrishnan, S.; Agrawal, A.; and Lee, S. 2018.
\newblock Overcoming language priors in visual question answering with
  adversarial regularization.
\newblock In \emph{NeurIPS}, 1541--1551.

\bibitem[{Ren et~al.(2019)Ren, Deng, He, and Che}]{ren2019generating}
Ren, S.; Deng, Y.; He, K.; and Che, W. 2019.
\newblock Generating natural language adversarial examples through probability
  weighted word saliency.
\newblock In \emph{Proceedings of the 57th Annual Meeting of the Association
  for Computational Linguistics}, 1085--1097.

\bibitem[{Ribeiro, Singh, and Guestrin(2018)}]{ribeiro2018semantically}
Ribeiro, M.~T.; Singh, S.; and Guestrin, C. 2018.
\newblock Semantically equivalent adversarial rules for debugging nlp models.
\newblock In \emph{ACL}, 856--865.

\bibitem[{Richardson, Burges, and Renshaw(2013)}]{richardson2013mctest}
Richardson, M.; Burges, C.~J.; and Renshaw, E. 2013.
\newblock Mctest: A challenge dataset for the open-domain machine comprehension
  of text.
\newblock In \emph{EMNLP}, 193--203.

\bibitem[{Roese(1997)}]{Roese1997cfthinking}
Roese, N. 1997.
\newblock Counterfactual thinking.
\newblock \emph{Psychological Bulletin} 121(1): 133--148.

\bibitem[{Rothman and Greenland(2005)}]{rothman2005causation}
Rothman, K.~J.; and Greenland, S. 2005.
\newblock Causation and causal inference in epidemiology.
\newblock \emph{American journal of public health} 95(S1): S144--S150.

\bibitem[{Shoeybi et~al.(2019)Shoeybi, Patwary, Puri, LeGresley, Casper, and
  Catanzaro}]{shoeybi2019megatron}
Shoeybi, M.; Patwary, M.; Puri, R.; LeGresley, P.; Casper, J.; and Catanzaro,
  B. 2019.
\newblock Megatron-lm: Training multi-billion parameter language models using
  gpu model parallelism.
\newblock \emph{arXiv preprint arXiv:1909.08053} .

\bibitem[{Steel(2004)}]{steel2004social}
Steel, D. 2004.
\newblock Social mechanisms and causal inference.
\newblock \emph{Philosophy of the social sciences} 34(1): 55--78.

\bibitem[{Sun et~al.(2019)Sun, Yu, Chen, Yu, Choi, and Cardie}]{sun2019dream}
Sun, K.; Yu, D.; Chen, J.; Yu, D.; Choi, Y.; and Cardie, C. 2019.
\newblock DREAM: A Challenge Data Set and Models for Dialogue-Based Reading
  Comprehension.
\newblock \emph{TACL} 7: 217--231.

\bibitem[{Szegedy et~al.(2013)Szegedy, Zaremba, Sutskever, Bruna, Erhan,
  Goodfellow, and Fergus}]{SzegedyZSBEGF13}
Szegedy, C.; Zaremba, W.; Sutskever, I.; Bruna, J.; Erhan, D.; Goodfellow, I.;
  and Fergus, R. 2013.
\newblock Intriguing properties of neural networks.
\newblock \emph{arXiv preprint arXiv:1312.6199} .

\bibitem[{Tang, Huang, and Zhang(2020)}]{tang2020long}
Tang, K.; Huang, J.; and Zhang, H. 2020.
\newblock Long-Tailed Classification by Keeping the Good and Removing the Bad
  Momentum Causal Effect.
\newblock \emph{arXiv preprint arXiv:2009.12991} .

\bibitem[{Tang et~al.(2020)Tang, Niu, Huang, Shi, and Zhang}]{tang2020unbiased}
Tang, K.; Niu, Y.; Huang, J.; Shi, J.; and Zhang, H. 2020.
\newblock Unbiased scene graph generation from biased training.
\newblock \emph{arXiv preprint arXiv:2002.11949} .

\bibitem[{Utama, Moosavi, and Gurevych(2020)}]{utama2020mind}
Utama, P.~A.; Moosavi, N.~S.; and Gurevych, I. 2020.
\newblock Mind the Trade-off: Debiasing NLU Models without Degrading the
  In-distribution Performance.
\newblock \emph{arXiv preprint arXiv:2005.00315} .

\bibitem[{Van~der Laan and Rose(2011)}]{van2011targeted}
Van~der Laan, M.~J.; and Rose, S. 2011.
\newblock \emph{Targeted learning: causal inference for observational and
  experimental data}.
\newblock Springer Science \& Business Media.

\bibitem[{Wallace et~al.(2019)Wallace, Feng, Kandpal, Gardner, and
  Singh}]{wallace2019universal}
Wallace, E.; Feng, S.; Kandpal, N.; Gardner, M.; and Singh, S. 2019.
\newblock Universal Adversarial Triggers for Attacking and Analyzing NLP.
\newblock In \emph{EMNLP}, 2153--2162.

\bibitem[{Wang et~al.(2020)Wang, Huang, Zhang, and Sun}]{wang2020visual}
Wang, T.; Huang, J.; Zhang, H.; and Sun, Q. 2020.
\newblock Visual commonsense r-cnn.
\newblock In \emph{CVPR}, 10760--10770.

\bibitem[{Wang and Bansal(2018)}]{wang2018robust}
Wang, Y.; and Bansal, M. 2018.
\newblock Robust Machine Comprehension Models via Adversarial Training.
\newblock In \emph{NAACL}, 575--581.

\bibitem[{Williams, Nangia, and Bowman(2018)}]{williams2018broad}
Williams, A.; Nangia, N.; and Bowman, S. 2018.
\newblock A Broad-Coverage Challenge Corpus for Sentence Understanding through
  Inference.
\newblock In \emph{Proceedings of the 2018 Conference of the North American
  Chapter of the Association for Computational Linguistics: Human Language
  Technologies, Volume 1 (Long Papers)}, 1112--1122.

\bibitem[{Wolf et~al.(2019)Wolf, Debut, Sanh, Chaumond, Delangue, Moi, Cistac,
  Rault, Louf, Funtowicz, and Brew}]{Wolf2019HuggingFacesTS}
Wolf, T.; Debut, L.; Sanh, V.; Chaumond, J.; Delangue, C.; Moi, A.; Cistac, P.;
  Rault, T.; Louf, R.; Funtowicz, M.; and Brew, J. 2019.
\newblock HuggingFace's Transformers: State-of-the-art Natural Language
  Processing.
\newblock \emph{arXiv preprint arXiv:1910.03771} .

\bibitem[{Yaghoobzadeh et~al.(2019)Yaghoobzadeh, Tachet, Hazen, and
  Sordoni}]{yaghoobzadeh2019robust}
Yaghoobzadeh, Y.; Tachet, R.; Hazen, T.~J.; and Sordoni, A. 2019.
\newblock Robust natural language inference models with example forgetting.
\newblock \emph{arXiv preprint arXiv:1911.03861} .

\bibitem[{Yang, Zhang, and Cai(2020)}]{yang2020deconfounded}
Yang, X.; Zhang, H.; and Cai, J. 2020.
\newblock Deconfounded image captioning: A causal retrospect.
\newblock \emph{arXiv preprint arXiv:2003.03923} .

\bibitem[{Yang et~al.(2019)Yang, Dai, Yang, Carbonell, Salakhutdinov, and
  Le}]{yang2019xlnet}
Yang, Z.; Dai, Z.; Yang, Y.; Carbonell, J.; Salakhutdinov, R.~R.; and Le, Q.~V.
  2019.
\newblock Xlnet: Generalized autoregressive pretraining for language
  understanding.
\newblock In \emph{NeurIPS}, 5754--5764.

\bibitem[{Yeh and Chen(2019)}]{yeh2019qainfomax}
Yeh, Y.-T.; and Chen, Y.-N. 2019.
\newblock QAInfomax: Learning Robust Question Answering System by Mutual
  Information Maximization.
\newblock In \emph{EMNLP}, 3361--3366.

\bibitem[{Yue et~al.(2020)Yue, Zhang, Sun, and Hua}]{yue2020interventional}
Yue, Z.; Zhang, H.; Sun, Q.; and Hua, X. 2020.
\newblock Interventional Few-Shot Learning.
\newblock \emph{arXiv preprint arXiv:2009.13000} .

\bibitem[{Zhang et~al.(2020{\natexlab{a}})Zhang, Zhang, Tang, Hua, and
  Sun}]{zhang2020causal}
Zhang, D.; Zhang, H.; Tang, J.; Hua, X.; and Sun, Q. 2020{\natexlab{a}}.
\newblock Causal Intervention for Weakly-Supervised Semantic Segmentation.
\newblock \emph{arXiv preprint arXiv:2009.12547} .

\bibitem[{Zhang et~al.(2019)Zhang, Zhou, Miao, and Li}]{zhang2019generating}
Zhang, H.; Zhou, H.; Miao, N.; and Li, L. 2019.
\newblock Generating Fluent Adversarial Examples for Natural Languages.
\newblock In \emph{Proceedings of the 57th Annual Meeting of the Association
  for Computational Linguistics}, 5564--5569.

\bibitem[{Zhang et~al.(2020{\natexlab{b}})Zhang, Sheng, Alhazmi, and
  Li}]{zhang2020adversarial}
Zhang, W.~E.; Sheng, Q.~Z.; Alhazmi, A.; and Li, C. 2020{\natexlab{b}}.
\newblock Adversarial Attacks on Deep-learning Models in Natural Language
  Processing: A Survey.
\newblock \emph{ACM Transactions on Intelligent Systems and Technology} 11(3):
  1--41.

\bibitem[{Zhang, Baldridge, and He(2019)}]{zhang2019paws}
Zhang, Y.; Baldridge, J.; and He, L. 2019.
\newblock PAWS: Paraphrase Adversaries from Word Scrambling.
\newblock In \emph{NAACL}, 1298--1308.

\end{thebibliography}
\newpage
\renewcommand\thesubsection{\Alph{subsection}}

\begin{center}
    \textbf{\Large Supplementary Materials}
\end{center}
These supplementary materials include the entire algorithm, the derivation of 
loss functions, the generation progress of \texttt{AddVerb}, more implementation details and more experiment results for ablation study, case study and data augmentation.

\section{Section A. Algorithm}
\label{CVC algorithm}
\begin{algorithm}[h]
\caption{Counterfactual Variable Control (CVC) algorithm}
\textbf{Stage one: multi-task training} \\
\textbf{Input:} complete train set data $\mathcal{X}$ and $N$ different subsets of train set data $\{\mathcal{X}_n\}^N_{n=1}$\\
\textbf{Output:} $F^r$ with parameters $\theta^r$ and $\{F^s_n\}^N_{n=1}$ with parameters $\{\theta^s_n\}^N_{n=1}$
\begin{algorithmic}[1]
\FOR{\texttt{batch} in $\mathcal{X}$ and $\{\mathcal{X}_n\}^N_{n=1}$}
\FOR{$n$ in $\{1,...,N\}$}
\STATE optimize $\theta^s_n$ with \texttt{batch} of $\mathcal{X}_n$ by Eq. 9;
\ENDFOR
\STATE optimize $\theta^r$ with \texttt{batch} of $\mathcal{X}$ by Eq. 11 for MCQA (by $\mathcal{L}^{e2}$ in Eq. 12 for SEQA);
\ENDFOR
\vspace*{0.05in}
\end{algorithmic}
\textbf{Stage two: counterfactual inference} \\
\textbf{Input:} $F^r$ with parameters $\theta^r$, $\{F^s_n\}^N_{n=1}$ with parameters $\{\theta^s_n\}^N_{n=1}$, complete target test data $\mathcal{X}^{\prime}$ along with its subsets $\{\mathcal{X}_n^{\prime}\}^N_{n=1}$ and a boolean $USE\_IV$ (if $USE\_IV=False$, train set data is also used)\\
\textbf{Output:} CVC-IV inference result or CVC-MV inference result ($\{F^c_n\}^N_{n=1}$ with parameters $\{\theta^c_n\}^N_{n=1}$)
\begin{algorithmic}[1]
\IF{$USE\_IV$}
\STATE compute CVC-IV inference result by Eq. 14;
\ELSE 
\STATE optimize $\{\theta^c_n\}^N_{n=1}$ with $\mathcal{X}$ and $\{\mathcal{X}_n\}^N_{n=1}$ by Eq. 15, Eq. 16 and cross-entropy loss for QA task;
\STATE compute CVC-MV inference result with target test data $\mathcal{X}^{\prime}$ and $\{\mathcal{X}_n^{\prime}\}^N_{n=1}$ by Eq. 15 and Eq. 16;
\ENDIF
\end{algorithmic}
\label{cvc-algorithm}
\end{algorithm}
\noindent{\color{blue}This is supplementary to Section ``The Implementation of CVC''}
In Algorithm~\ref{cvc-algorithm}, we summarize the overall process of the proposed Counterfactual Variable Control (CVC) approach. The process consists of two stages: multi-task training (Section 3.1) and counterfactual inference (Section 3.2).
Multi-task training aims to train a robust branch $F^r$ and $N$ shortcut branches $\{F^s_n\}^N_{n=1}$.
Counterfactual inference performs the robust and interpretable reasoning for QA.
$USE\_IV\!=\!True$ means using CVC-IV inference and $USE\_IV\!=\!False$ means using CVC-MV inference.
Line 4 in counterfactual inference can be elaborated as three steps: (1) it uses $c$-adaptor to compute $c_{n}^r$ according to Eq. 16 firstly; (2) it derives the logits of CVC-MV inference based on Eq. 15; and (3) it updates $c$-adaptor using cross-entropy loss (computed between the CVC-MV logits and the ground truth). Line 4 aims to train the $c$-adaptor with train set data.

\section{Section B. Loss Functions}
\label{Loss function}
{\color{blue}This is supplementary to Eq. 11 and Eq. 12.}
Eq. 11 is a straightforward version of loss according to Eq. 10.
Here, we prove that Eq. 12 has the same objective as Eq. 11.
Assume $j$ is the index of the ground-truth, 
the equation expansion to Eq. 11 is shown as follows:
\begin{equation}
\begin{aligned}
    \mathcal{L}^e
    &=-\sum_i{\mathtt{p}_i\log\softmax\big(\sum_n{\hat{\mathtt{p}}_{i}^r\cdot \hat{\mathtt{p}}_{n,i}^s}\big)}\\
    &=-\log\softmax\big(\sum_n{\hat{\mathtt{p}}_{j}^r\cdot \hat{\mathtt{p}}_{n,j}^s}\big) \\
    &=-\log\frac{e^{\sum_n{\hat{\mathtt{p}}_{j}^r\cdot \hat{\mathtt{p}}_{n,j}^s}}}{\sum_i{e^{\sum_n{\hat{\mathtt{p}}_{i}^r\cdot \hat{\mathtt{p}}_{n,i}^s}}}}\\
    &=\log{\sum_i{e^{\sum_n{\hat{\mathtt{p}}_{i}^r\cdot \hat{\mathtt{p}}_{n,i}^s}}}}-\sum_n{\hat{\mathtt{p}}_{j}^r\cdot \hat{\mathtt{p}}_{n,j}^s}
\end{aligned}
\label{le_expand}
\end{equation}
Similarly, the equation expansion to Eq. 12 (we use $\mathcal{L}^{e1}$ here since each element in $w_n$ is always positive and we are not able to access to $w_n$ during inference stage) is shown as:
\begin{equation}
    \begin{aligned}
        \mathcal{L}^{e1}&=-\sum_n\sum_i{\mathtt{p}_i\log\mathrm{softmax}(\hat{\mathtt{p}}_{i}^r\cdot\hat{\mathtt{p}}_{n,i}^s)}\\
        &=-\sum_n\log\softmax(\hat{\mathtt{p}}_{j}^r\cdot\hat{\mathtt{p}}_{n,j}^s)\\
        &=-\sum_n\log\frac{e^{\hat{\mathtt{p}}_{j}^r\cdot\hat{\mathtt{p}}_{n,j}^s}}{\sum_i{e^{\hat{\mathtt{p}}_{i}^r\cdot\hat{\mathtt{p}}_{n,i}^s}}}\\
        &=-\sum_n\big({\hat{\mathtt{p}}_{j}^r\cdot \hat{\mathtt{p}}_{n,j}^s}-\log\sum_i{e^{\hat{\mathtt{p}}_{i}^r\cdot\hat{\mathtt{p}}_{n,i}^s}}\big)\\
        &=\sum_n\log\sum_i{e^{\hat{\mathtt{p}}_{i}^r\cdot\hat{\mathtt{p}}_{n,i}^s}}-\sum_n{\hat{\mathtt{p}}_{j}^r\cdot \hat{\mathtt{p}}_{n,j}^s}
    \end{aligned}
\label{le1_expand}
\end{equation}
From above equations, we see that both Eq. 11 and Eq. 12 aim to maximize $\sum_n{\hat{\mathtt{p}}_{j}^r\cdot \hat{\mathtt{p}}_{n,j}^s}$, i.e., the ``adjusting'' function in Eq. 10.

\section{Section C. AddVerb}
\label{AddVerb}
\begin{figure}[h]
    \centering
    \includegraphics[width=0.48\textwidth]{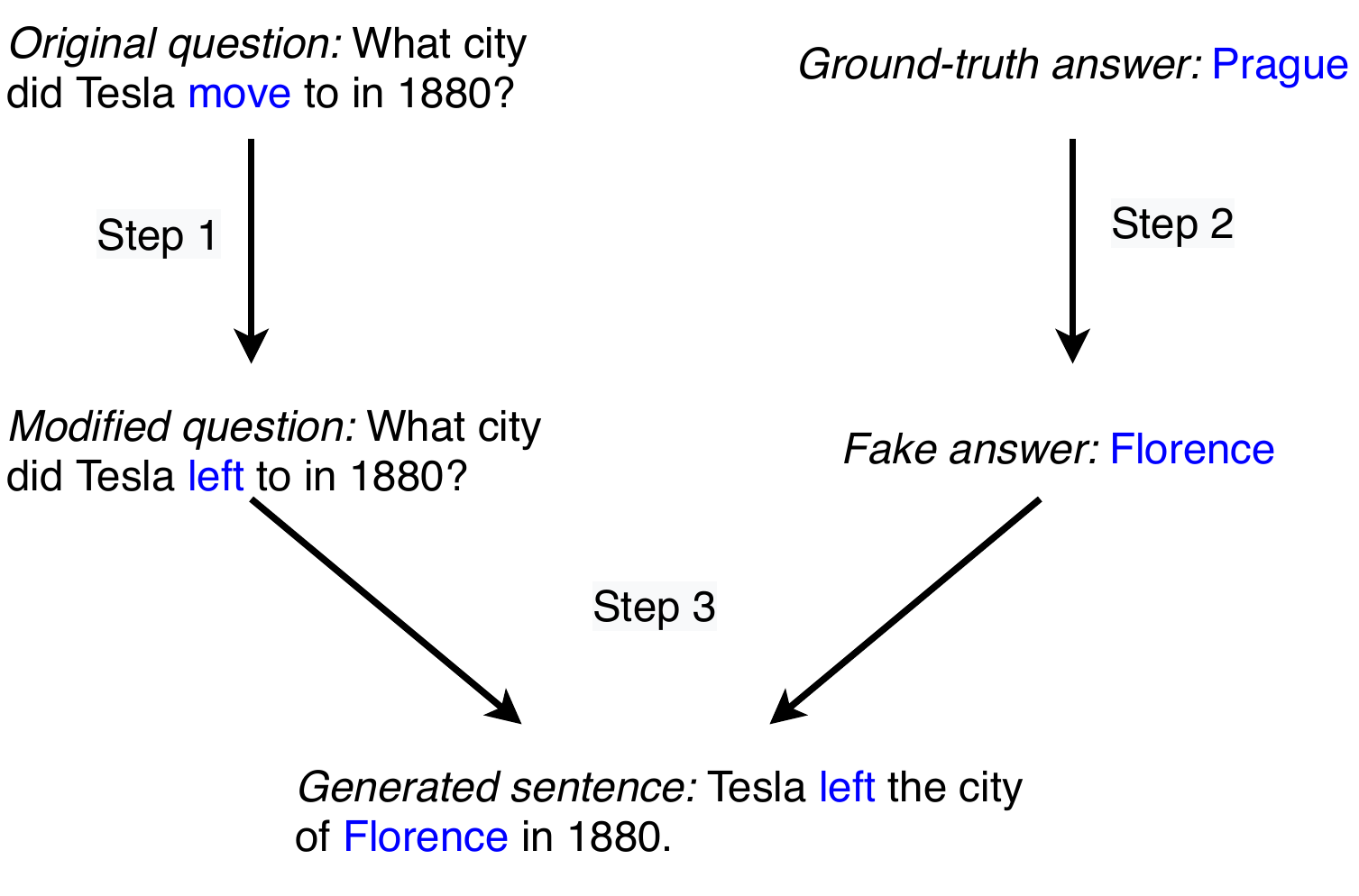}
    \caption{An illustration of the \texttt{AddVerb}. We use the instanced question-answer pair in~\cite{jia2017adversarial} as an example.}
    \label{fig:addverb}
\end{figure}

{\color{blue}This is supplementary to the \texttt{AddVerb} in ``Adversarial Attack on SEQA''.}
The generation process of \texttt{AddVerb} is similar to that of \texttt{AddSent}~\cite{jia2017adversarial}. The differences consists of (1) \texttt{AddVerb} is used to evaluate the robustness of the model against verb attacks and (2) \texttt{AddVerb} instances are annotated by a human expert linguist completely (raw version of \texttt{AddSent} is firstly generated by machine).
Given a question-answer pair, the linguist creates a distracting \texttt{AddVerb} sentence in three steps:
\begin{itemize}
    \item Replace the verb in the question with an antonym of this verb or an irrelevant verb.
    \item Create a fake answer with the same type as the ground-truth answer.
    \item Combine modified question and fake answer, and convert them into the statement.
\end{itemize}
An illustration of the whole process is shown in Figure~\ref{fig:addverb}.

\section{Section D. Implementation Details}
\label{Implementation details}

\subsection{General Implementation}
{\color{blue}This is supplementary to ``Implementation Details'' in Section ``Experiments''.} The overall experiments are conducted on two pieces of Tesla V100 or two pieces of RTX 2080Ti (depending on the useage of memory).

\noindent\textbf{Multi-task Training.}
The learning rates are fixed to 3e-5 and 2e-5 for BERT-base and BERT-large, respectively.
The number of bottom shared layers is fixed to $5/6$ of the total number of layers in the backbone language model for parameter-efficiency, \eg, sharing 10 layers in bottom shared layers when the BERT-base (12 layers) is adopted as the backbone.
A maximum sequence length of 384 is adopted in text preprocessing,
tokens out of the maximum sequence length are truncated and sequences shorter than 384 will be padded.
The number of training epochs is selected amongst $\{16,24,36\}$. 
The batch size is selected amongst $\{16,24,32\}$. 
A linear warm-up strategy for learning rates is used with the first $10\%$ steps in the whole multi-branch training stage.
Gradient accumulation and half precision are used to relieve the issue of huge memory usage. 

\noindent\textbf{Counterfactual Inference.}
We use CVC-IV inference for model selection, and apply the chosen checkpoint in both CVC-IV and CVC-MV.
The number of epochs in $c$-adaptor training is selected amongst $\{1,3,6,10\}$ and the batch size of $c$-adaptor training is fixed to $24$. Same learning rates and same warm-up strategy for learning rates in multi-task training are used in $c$-adaptor training.

\subsection{MCQA-Specific}
\begin{table}[h]
\centering
\begin{tabular}{llll}
 \toprule
                 & MCTest & DREAM & RACE \\                \midrule
Random guess     & 25.0   & 33.3  & 25.0 \\
Complete input & 68.9   & 61.5  & 64.7 \\
No \textit{P}   & 24.2   & 32.8  & 41.6 \\
No \textit{Q}   & 52.5   & 57.1  & 51.0 \\
No \textit{P}, \textit{Q}     & 22.4   & 33.4  & 34.7 \\\bottomrule
\end{tabular}
\caption{Accuracies (\%) of BERT-base MCQA models trained with complete input. ``No $X$'' means the value of input variable $X$ is muted.}
\label{remove_part_mcqa}
\end{table}
\noindent{\color{blue}This is supplementary to ``MCQA-Specific'' in Section ``Experiments''.} 
The results of muting experiments on three MCQA datasets are shown in Table~\ref{remove_part_mcqa}.
Each number in this table indicates the strength of corresponding direct cause-effects.
For example, the results on the row of 
``No \textit{Q}'' represent the effects of $\textit{P}\!\rightarrow\!\textit{A}$ and $\textit{O}\!\rightarrow\!\textit{A}$ shown in Figure 2~(b). 
It is clearly shown that the effect of $\textit{Q}\!\rightarrow\!\textit{A}$ is negligible compared to that of $\textit{P}\!\rightarrow\!\textit{A}$.
%
Therefore, we ignore $\textit{Q}\!\rightarrow\!\textit{A}$ when implementing shortcut branches for MCQA.
We use one shortcut branch ($N=1$) with input $\mathcal{X}_1=\{\textit{P},\textit{O}\}$, to learn $\textit{P}\!\rightarrow\!\textit{A}$ and $\textit{O}\!\rightarrow\!\textit{A}$. Other MCQA-specific implementation details (\eg, how to design the FC layer) are the same with the official code of~\cite{devlin2019bert}.

\subsection{SEQA-specific}
\begin{figure}[h]
    \centering
    \includegraphics[width=0.5\textwidth]{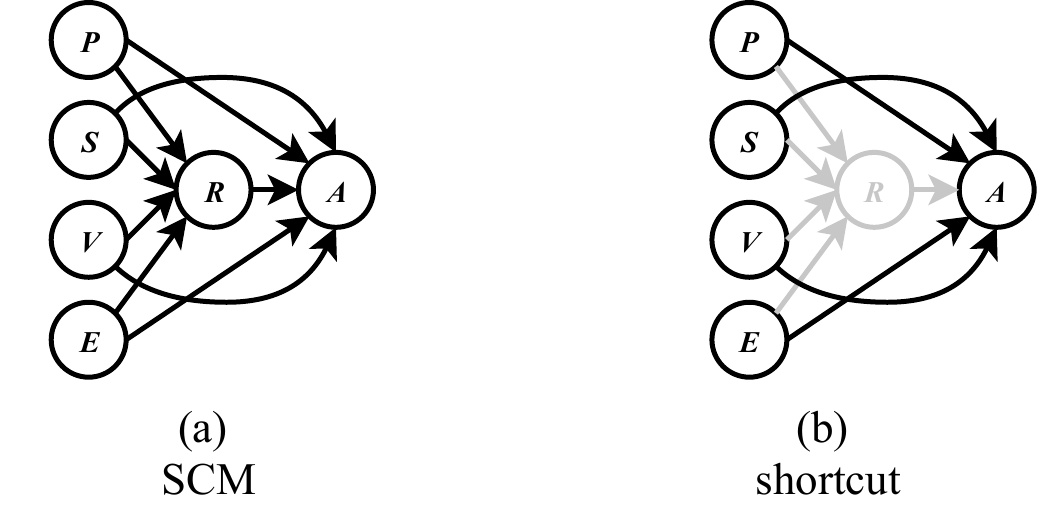}
    \caption{The SCM for SEQA task where \textit{Q} is decomposed to \textit{S}, \textit{V} and \textit{E}.}
    \label{fig_human_machine_2}
\end{figure}
\begin{table}[h]
\centering
\begin{tabular}{ll}
\toprule
   & SQuAD \\ \midrule
Complete input & 88.1  \\
No \textit{E} & 59.4  \\
No \textit{V} & 55.1  \\
No \textit{E}, \textit{V} & 15.3 \\
No \textit{Q} & 12.4 \\\bottomrule
\end{tabular}
\caption{F1 scores (\%) of BERT-base SEQA models trained with complete input. ``No $X$'' means the value of input variable $X$ is muted.}
\label{remove_part_seqa}
\end{table}
{\color{blue}This is supplementary to ``SEQA-Specific'' in Section ``Experiments''.} 

\noindent\textbf{Partition Details.}
As mentioned in the main paper, we partition the \textit{Q} in SEQA into \textit{E}, \textit{V} and \textit{S}. 
The SCM for SEQA is shown in Figure~\ref{fig_human_machine_2}. 
\begin{itemize}
    \item For extracting \textit{E} in \textit{Q}, we utilize Stanford part-of-speech tagger and collect the words with the following labels as \textit{E}: $\{\text{NN}, \text{NNS}, \text{NNP}, \text{NNPS}, \text{CD}, \text{SYM}, \text{FW}\}$.
    \item For extracting \textit{S} in \textit{Q}, we firstly retrieve the stop words from NLTK and all punctuations, and then remove words of negation such as ``don't'' and interrogative words such as ``where''. We expect to gather less important tokens into \textit{S} (with negligible semantic meanings).
    \item We group the rest of the words, \eg, verbs and adverbs, and denote them as \textit{V}.
\end{itemize}
We show the results of muted experiments on SEQA dataset (SQuAD) in Table~\ref{remove_part_seqa}.

\noindent\textbf{Why Conduct This Partition?}
The reason is simple: \textit{P} is mandatory for SEQA.
Without \textit{P} will result in a null prediction of the model.
To study the effects of $\textit{Q}\!\rightarrow\!\textit{A}$, what we can do is to split the variable \textit{Q} into partitions.
%
Besides, our resulting \textit{Q} partitions are intuitive. \textit{E} and \textit{V} contain the most important semantic meanings. 

\noindent\textbf{Shortcut Branches.}
We adopt two shortcut branches ($N=2$) to represent all shortcut paths in Figure~\ref{fig_human_machine_2}(b).
The first shortcut branch takes $\mathcal{X}_1=\{\textit{P},\textit{S},\textit{V}\}$ as input and aims to learn $\textit{P}\!\rightarrow\!\textit{A}$, $\textit{S}\!\rightarrow\!\textit{A}$ and $\textit{V}\!\rightarrow\!\textit{A}$.
The second shortcut branch takes $\mathcal{X}_1=\{\textit{P},\textit{S},\textit{E}\}$ as input and learns $\textit{P}\!\rightarrow\!\textit{A}$, $\textit{S}\!\rightarrow\!\textit{A}$ and $\textit{E}\!\rightarrow\!\textit{A}$. 
Other SEQA-specific implementation details (\eg, how to deploy answer pointer layer to make prediction) are the same with the official code of~\cite{devlin2019bert}.

\section{Section E. More Experimental Results}
\label{Experiment results}
\subsection{Overall Results Compared to Baselines and State-of-the-arts.}
{\color{blue}This is supplementary to ``Overall Results Compared to Baselines and State-of-the-arts.'' in Section ``Experiments''.} We show the full result and implementation details of Table 4 in Table~\ref{table:main_results_aug_MCQA} and Table~\ref{table:main_results_aug_SEQA}.
\begin{table*}[t]
\centering
\begin{tabular}{llllllllllllll}
\toprule                                            
Dataset & Method    & \texttt{Test}                      & \texttt{Adv1}                      & \texttt{Adv2}                      & \texttt{Adv3}                      & \texttt{Adv4}                      & A.G.                             \\ \midrule
\multicolumn{1}{l}{\multirow{6}{*}{\textbf{MCTest}}} & \multicolumn{1}{l}{CT~\cite{devlin2019bert}} & \multicolumn{1}{l}{68.9} & \multicolumn{1}{l}{63.9} & \multicolumn{1}{l}{59.4} & \multicolumn{1}{l}{20.2} & \multicolumn{1}{l}{54.8} & \multicolumn{1}{l}{-}  \\ 
\multicolumn{1}{c}{} & DRiFt~\cite{he2019unlearn} & 69.6  & 66.0  & 61.9 & 23.0 & 54.8 & +1.9\% \\
\multicolumn{1}{c}{} & Bias Product~\cite{clark2019don} & 71.0  & 66.7  & 63.6 & 22.8 & 65.5 & +5.1\% \\
\multicolumn{1}{c}{} & Learned-Mixin~\cite{clark2019don} & 70.5  & 66.2  & 60.4 & 20.2 & 58.8 & +1.8\% \\
\multicolumn{1}{c}{}                        & CVC-MV                      & 68.1                     & 69.1                      & 65.6                     & 26.8                      & 61.0         & +6.1\%            \\
\multicolumn{1}{c}{}                        & CVC-IV                           & 69.4                      & 70.0                        & 65.4                      & 28.7                      & 59.9        & \textbf{+6.4\%}         \\
 \midrule
\multicolumn{1}{l}{\multirow{6}{*}{\textbf{DREAM}}}  & \multicolumn{1}{l}{CT~\cite{devlin2019bert}} & \multicolumn{1}{l}{61.5} & \multicolumn{1}{l}{47.5} & \multicolumn{1}{l}{39.2} & \multicolumn{1}{l}{20.9} & \multicolumn{1}{l}{41.8} & \multicolumn{1}{l}{-} \\ 
\multicolumn{1}{c}{} & DRiFt~\cite{he2019unlearn} & 60.1  & 48.5  & 42.2 & 23.9 & 44.7 & +2.5\% \\
\multicolumn{1}{c}{} & Bias Product~\cite{clark2019don} & 58.6  & 47.5  & 38.8 & 22.6 & 40.2 & -0.1\% \\
\multicolumn{1}{c}{} & Learned-Mixin~\cite{clark2019don} & 60.9  & 49.2  & 41.7 & 20.0 & 42.3 & +1.0\% \\
\multicolumn{1}{l}{}                        & CVC-MV                      & 60.1                      & 49.6                      & 39.9                      & 23.7                      & 45.6        & +2.3\%       
\\& CVC-IV                           & 60.0                        & 49.2                      & 40.7                      & 25.0                        & 47.1       & \textbf{+3.1\%}                           \\
                \midrule
\multicolumn{1}{l}{\multirow{6}{*}{\textbf{RACE}}}   & \multicolumn{1}{l}{CT~\cite{devlin2019bert}} & \multicolumn{1}{l}{64.7} & \multicolumn{1}{l}{56.0}   & \multicolumn{1}{l}{50.1} & \multicolumn{1}{l}{36.6} & \multicolumn{1}{l}{58.3} & \multicolumn{1}{l}{-} \\ 
\multicolumn{1}{c}{} & DRiFt~\cite{he2019unlearn} & 62.0  & 56.1  & 53.3 & 39.3 & 58.3 & \textbf{+1.7\%} \\
\multicolumn{1}{c}{} & Bias Product~\cite{clark2019don} & 62.3  & 56.7  & 53.3 & 37.0 & 56.8 & +1.0\% \\
\multicolumn{1}{c}{} & Learned-Mixin~\cite{clark2019don} & 64.3  & 56.5  & 51.9 & 38.0 & 60.1 & +1.4\% \\
\multicolumn{1}{l}{}                        & CVC-MV                      & 64.4                          & 56.7                          & 51.7                          & 39.1                          & 59.2         & +1.4\%                  \\
& CVC-IV                           & 64.1                      & 57.0                        & 52.2                      & 38.8                      & 58.6   & +1.4\%                                        \\
 \bottomrule 
\end{tabular}
\caption{
MCQA accuracies (\%) on three datasets.
Models are trained on the original training data with BERT-base.
``A.G.'': our average improvement over the conventional training method (CT)~\cite{devlin2019bert} for \texttt{Adv*}.}
\label{table:main_results_aug_MCQA}
\end{table*}
\begin{table*}[h]
\centering
\begin{tabular}{llllll}
\toprule 
 Method  & \texttt{Dev}   & \texttt{Adv1}  & \texttt{Adv2}  & \texttt{Adv3}  & A.G.  \\
\midrule 
CT~\cite{devlin2019bert}          & 88.4  & 49.9  & 59.7  & 44.6 & -\\
DRiFt~\cite{he2019unlearn} & 85.7  & 53.7  & 65.7 & 48.5 & +4.5\% \\
Bias Product~\cite{clark2019don} & 87.8  & 53.6  & 65.7 & 47.3 & +4.1\% \\
Learned-Mixin~\cite{clark2019don} & 87.2  & 53.1  & 63.9 & 45.5  & +2.1\% \\
QAInformax~\cite{yeh2019qainfomax} & 88.6  & 54.5  & 64.9  & -   & +4.9\%  \\
CVC-MV   & 87.2  & 55.7  & 65.3  & 51.3 & +6.0\% \\
CVC-IV      & 86.6  & 56.3  & 66.2  & 51.5 & \textbf{+6.6\%} \\
\bottomrule 
\end{tabular}
\caption{SEQA F1-measure (\%) on the SQuAD \texttt{Dev} set (as \texttt{Test} set is not public) and adversarial sets.
Models are trained on the original training data with BERT-base.
%
``A.G.'': our average improvement over the conventional training method (CT)~\cite{devlin2019bert} for \texttt{Adv*}.}
\label{table:main_results_aug_SEQA}
\end{table*}

On MCQA task,  we implement DRiFt~\cite{he2019unlearn}, Bias Product~\cite{clark2019don} and Learned-Mixin~\cite{clark2019don} following ours CVC implementation (the design of shortcut branch) with their adjustment functions. 

On SEQA task, we implement DRiFt by directly changing our adjustment function (Eq. 10) to its. For Bias Product and Learned-Mixin, the corresponding adjustment functions in~\cite{clark2019don} are used, we also use TF-IDF as the shortcut branch in our experiment which is described in its paper.

From the two tables we can observe that CVC outperforms other methods in most of the cases. 

\subsection{Ablation Study}
{\color{blue}This is supplementary to ``Ablation Study'' in Section ``Experiments''.}
We conduct the ablation study for MCQA on MCTest with BERT-base.
In Table~\ref{ablation_mcqa}, we show the results obtained in 10 ablative settings. Specifically, we (1) use $\mathcal{X}_1=\{\textit{Q},\textit{O}\}$ as the input of the only shortcut branch;
(2) use two shortcut branches, where the first one takes $\mathcal{X}_1=\{\textit{P},\textit{O}\}$ as input and the second one takes $\mathcal{X}_2=\{\textit{Q},\textit{O}\}$ as input, and deploy the $\mathcal{L}^e$ in Eq. 11;
(3) use the same two shortcut branches as (2), but deploy the $\mathcal{L}^{e1}$ in Eq. 12;
(4) use the same two shortcut branches as (2), but $\mathcal{L}^{e2}$ in Eq. 12 is used;
(5) set $c_n^r$ to the same constant (tuned in $\{0.2,0.4,0.6,0.8,1\}$) for all input samples;
(6) let $c_n^r=\mathbf{JS}[\hat{p}^{r}||\hat{p}_{n}^s]$;
(7) let $c_n^r=\left|\hat{p}^{r}-\hat{p}_{n}^s\right|^2/2$ where $2$ is the upper bound of the square of Euclidean distance;
(8) remove the $distance$ item in Eq. 16;
(9) remove $\hat{p}^{r}$ and $\hat{p}_{n}^s$ in Eq. 16;
and (10) use the features extracted via top layers as input in Eq. 16.
\begin{table*}[h]
\centering
\begin{tabular}{lllllll}
\toprule 
Ablative Setting & \texttt{Dev}  & \texttt{Adv1} & \texttt{Adv2} & \texttt{Adv3} & \texttt{Adv4} & Average \\\midrule
(1) one modified Shct.br. & 68.3 & 63.1 & 58.0 & 24.8 & 56.5  & 50.6\\
(2) two Shct.br. with $\mathcal{L}^e$  & 70.1 & 66.8 & 61.0 & 24.6 & 57.1 & 52.4\\
(3) two Shct.br. with $\mathcal{L}^{e1}$   & 70.2 & 66.7 & 62.1 & 25.6 & 56.5 & 52.7\\
(4) two Shct.br. with $\mathcal{L}^{e2}$        & 70.8 & 66.6 & 61.8 & 27.1 & 62.2 & 54.4\\ 
\midrule
CVC-IV (ours)        & 69.4 & 70.0 & 65.4 & 28.7 & 59.9 & \textbf{56.0} \\ 
\midrule[0.8pt]
(5) same $c_{r,n}$   & 68.1 & 69.3 & 64.4 & 25.6  & 59.3 & 54.7\\ 
(6) $c_{r,n}\!=\!JS$ & 70.1 & 67.0 & 61.9 & 20.8 & 62.2 & 53.0\\
(7) $c_{r,n}\!=\!Euc$ & 69.8 & 67.7 & 61.9 & 22.3 & 60.5 & 53.1\\
(8) w/o $distance$  & 66.1 & 67.9 & 65.2 & 27.8 & 61.0 & 55.5\\
(9) w/o $\hat{p}_{r}$ and $\hat{p}_{n}$ & 65.6 & 66.3 & 64.8 & 27.4 & 59.9 & 54.6\\
(10) features as input & 68.6 & 68.2 & 62.9  & 23.4 & 59.9 & 53.6\\
\midrule
CVC-MV (ours)    & 68.1 & 69.1 & 65.6 & 26.8 & 61.0 & \textbf{55.6}\\ \bottomrule 
\end{tabular}
\caption{The ablation study on MCTest (BERT-base). (1)-(4) are ablative settings for multi-task training (using CVC-IV inference). ``Average'' means the average performance on \texttt{Adv*} test sets;
(5)-(10) are ablative settings related to CVC-MV inference.}
\label{ablation_mcqa}
\end{table*}

In Table~\ref{ablation_mcqa}, results on (1)-(4) show that considering the shortcut branch with input $\{\textit{Q,\textit{O}}\}$ 
is not effective for the robustness of model. 
The reason is that this shortcut branch is hard to train, \ie not easy to converge (please refer to ``MCQA-specific'').
%
The results suggest:
firstly, the shortcut branch with negligible effect magnitude can be ignored (when designing the networks);
secondly, if no prior knowledge of the effect magnitude on each shortcut path (of SCM), using $\mathcal{L}^{e2}$ is the best choice.
Results on (5)-(10) show the efficiency of our proposed $c$-adaptor. 

\subsection{Case study}
\begin{figure*}[h]
    \centering
    \includegraphics[width=1\textwidth]{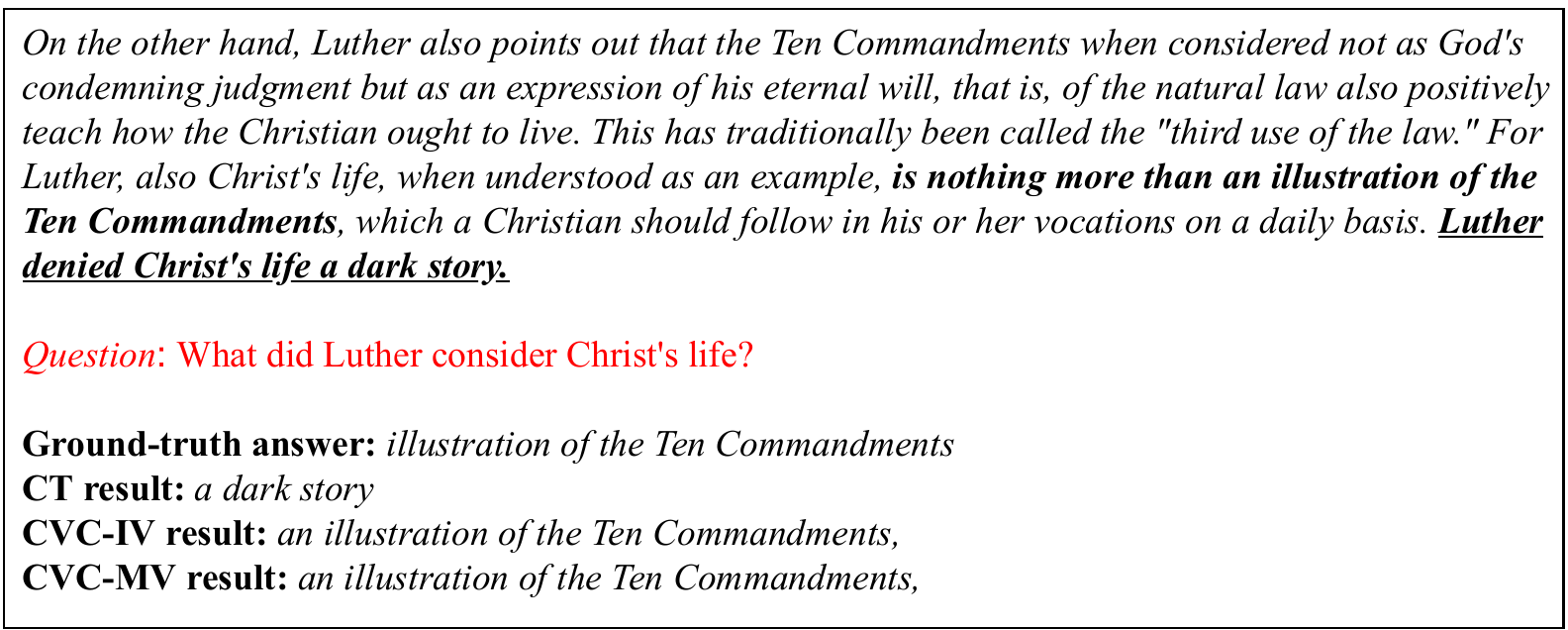}
    \caption{
    A case study of CVC on SQuAD trained on ofﬁcial data.
    The distracting sentence from \texttt{AddVerb} is \underline{underlined}.}
    \label{case_seqa}
\end{figure*}

{\color{blue}This is supplementary to ``Case Study'' in Section ``Experiments''.}
We visualize the respective examples of CVC-IV and CVC-MV inferences for SEQA. We show the contents of passage, question and predictions for CT, CVC-IV, and CVC-MV in Figure~\ref{case_seqa}.

\begin{figure*}[h]
    \centering
    \includegraphics[width=\textwidth]{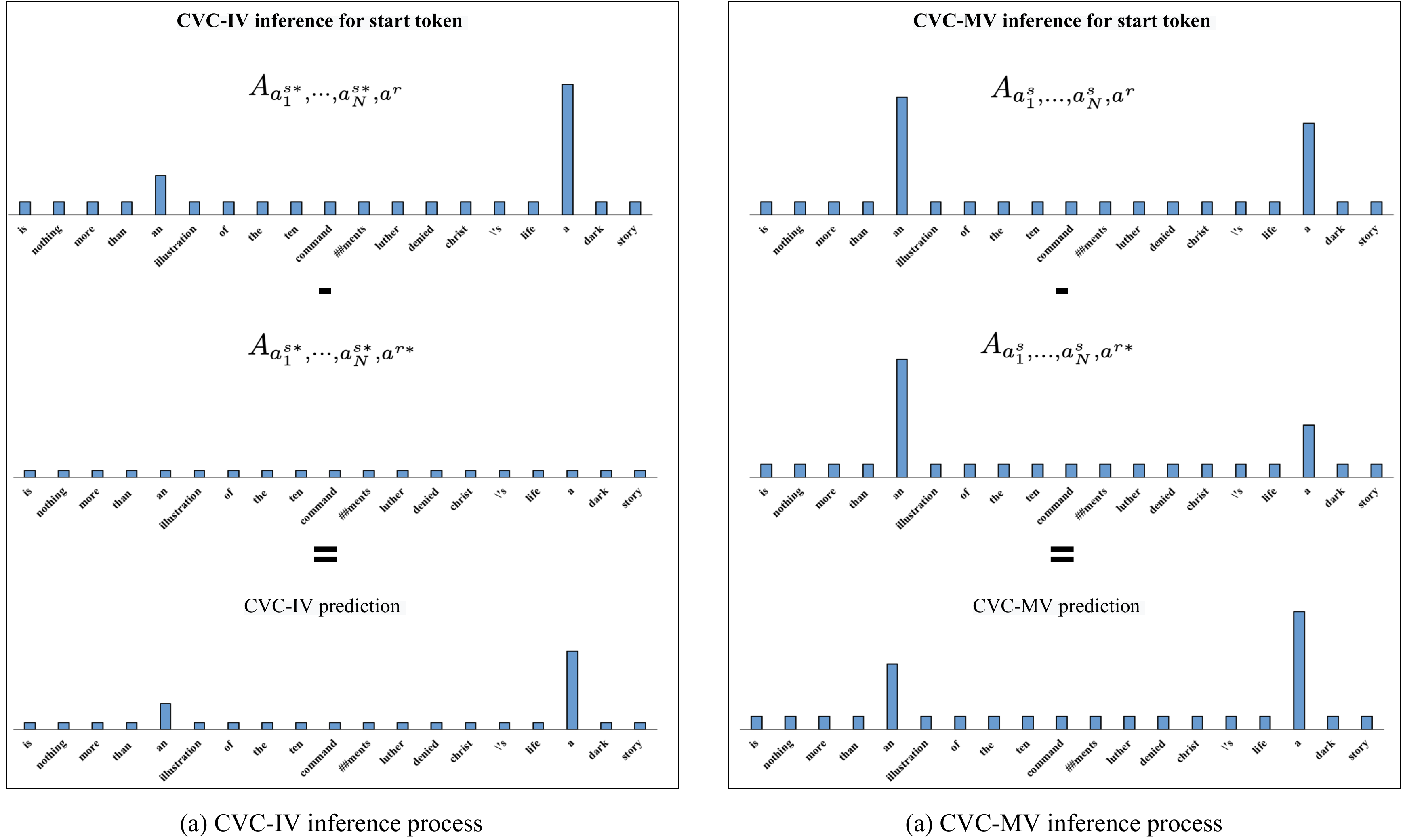}
    \caption{Process of CVC-IV inference and CVC-MV inference corresponding to the case in Figure~\ref{case_seqa}. Only bold tokens in passage are shown, due to limited page size. 
    }
    \label{case_seqa_cvcivmv}
\end{figure*}

Specifically, we illustrate CVC-IV and CVC-MV processes for predicting the start token in Figure~\ref{case_seqa_cvcivmv}, for simplicity.
We note that the process of predicting the end tokens is similar.

\subsection{Data augmentation}
{\color{blue}This is supplementary to ``Overall Results Compared to Baselines and State of theArt.'' in Section ``Experiments''.} 
An intuitive method to improve the model robustness is to augment the training data using adversarial examples~\cite{ribeiro2018semantically,jia2017adversarial}.
We conduct this experiment and show the results of MCTest
and present the results of three methods (CT, CVC-IV, and CVC-MV) in Table~\ref{data_aug_full}. 
Comparing Table~\ref{data_aug_full} to the result without data augmentation in main paper, we can see that models get consistently improved via data augmentation.
Comparing the results between CT and CVC-*, we find that the latter can achieve further performance boosts for augmented models, \eg, CVC-MV brings an average accuracy of $4\%$ to ``Add All'' models whose training data are augmented with four kinds of adversarial examples.
Please note that we skip the data augmentation experiments for SEQA, because the adversarial attacks (for SEQA) used in our work require a lot of human annotation and proofreading which are costly.

\begin{table*}[h]
\centering
\begin{tabular}{llllllllllll}
\toprule 
                                              &          & \texttt{Test} & \texttt{Adv1} & \texttt{Adv2} & \texttt{Adv3} & \texttt{Adv4} & AG \\\midrule
\multicolumn{1}{c}{\multirow{3}{*}{\textbf{Add Adv1}}} & CT       & 71.0 & 70.6 & 72.1 & 42.5 & 60.5 & - \\
\multicolumn{1}{c}{}                          & CVC-IV    & 71.7 & 73.3 & 74.9 & 49.2 & 63.8 & +3.9\% \\
\multicolumn{1}{c}{}                          & CVC-MV    & 71.6 & 72.9 & 74.8 & 48.0 & 62.7 & +3.2\% \\
\midrule
\multicolumn{1}{c}{\multirow{3}{*}{\textbf{Add Adv2}}} & CT       & 72.3 & 73.0 & 75.1 & 50.1 & 63.3 & - \\
\multicolumn{1}{c}{}                          & CVC-IV & 71.8 & 73.8 & 76.2 & 59.8 & 65.5 & +3.5\% \\
\multicolumn{1}{c}{}                          & CVC-MV & 71.8 & 74.2 & 76.6 & 61.1 & 65.5 & +3.9\% \\\midrule
\multirow{3}{*}{\textbf{Add Adv3}}                     & CT       & 67.5 & 62.7 & 59.9 & 70.9 & 57.1 & - \\
& CVC-IV    & 67.6 & 64.5 & 62.4 & 70.2 & 61.6 & +2.0\% \\
& CVC-MV & 66.8 & 63.7 & 62.3 & 70.3 & 60.5 & +1.5\% \\                
\midrule
\multirow{3}{*}{\textbf{Add Adv4}}                     & CT       & 69.8 & 65.4 & 60.2 & 27.7 & 63.3 & - \\
& CVC-IV    & 69.9 & 66.2 & 62.4 & 32.7 & 61.0 & +1.4\% \\
& CVC-MV    & 67.5 & 65.6 & 62.4 & 25.4 & 66.7 & +0.9\% \\ 
\midrule
\multirow{3}{*}{\textbf{Add All}}                      & CT       & 70.5 & 72.1 & 74.1 & 72.5 & 63.4 & - \\
& CVC-IV & 72.7 & 73.5 & 76.4 & 71.9 & 68.4 & +2.0\% \\ 
                                              & CVC-MV & \textbf{73.1} & \textbf{74.6} & \textbf{76.6} & \textbf{73.3} & \textbf{73.5} & \textbf{+4.0\%}\\
\bottomrule  
\end{tabular}
\caption{Accuracies (\%) on the MCTest dataset, using different kinds of data augmentation in training with BERT-base.}
\label{data_aug_full}
\end{table*}

\subsection{Extend our method to Natural Language Inference (NLI) task}
{\color{blue}We extend our method to NLI task and compare with other state-of-the-arts methods.}
\begin{table*}[h]
\centering
\begin{tabular}{lll}
\toprule
                   & Matched Dev & HANS \\\midrule
CT                 & 84.2        & 62.4 \\
Reweight~\cite{clark2019don}           & 83.5        & 69.2 \\
Bias Product~\cite{clark2019don}       & 83.0        & 67.9 \\
Learned-Mixin~\cite{clark2019don}      & 84.3        & 64.0 \\
Learned-Mixin+H~\cite{clark2019don}   & 84.0        & 66.2 \\
DRiFt-HYPO~\cite{he2019unlearn}         & 84.3        & 67.1 \\
DRiFt-HAND~\cite{he2019unlearn}         & 81.7        & 68.7 \\
DRiFt-CBOW~\cite{he2019unlearn}         & 82.1        & 65.4 \\
Mind the Trade-off~\cite{utama2020mind} & 84.3        & 70.3 \\
CVC-IV             & 82.9        & 70.0 \\
CVC-MV             & 83.0        & \textbf{71.5} \\\bottomrule
\end{tabular}
\caption{
NLI accuracies (\%) on Matched Dev and HANS.
Models are trained on the original training data with BERT-base.}
\label{table:main_results_NLI}
\end{table*}
Following the setting in previous works, we train the model on MNLI~\cite{williams2018broad} and evaluate on HANS~\cite{mccoy2019right}. We use the overlapped tokens in hypothesis and premise as the only bias branch in implementation of CVC.
The result is shown in Table~\ref{table:main_results_NLI}.

\end{document}